\documentclass[review]{elsarticle}

\usepackage{lineno,hyperref}
\modulolinenumbers[5]
\usepackage{algorithm}
\usepackage{algorithmic}

\usepackage{times}
\usepackage{epsfig}
\usepackage{graphicx}
\usepackage{amsmath}
\usepackage{amssymb}

\journal{Neurocomputing}

\begin{document}

\begin{frontmatter}

\title{Improving Variational Autoencoder with Deep Feature Consistent and Generative Adversarial Training}


\author[mymainaddress,myfifthhaddress]{Xianxu Hou}
\author[myfourthaddress]{Ke Sun}
\author[myfifthhaddress,mysecondaryaddress]{Linlin Shen}


\author[mymainaddress,myfifthhaddress,mythirdaddress]{Guoping Qiu\corref{mycorrespondingauthor}}
\cortext[mycorrespondingauthor]{Corresponding author}
\ead{qiu@szu.edu.cn, guoping.qiu@nottingham.ac.uk}

\address[mymainaddress]{College of Information Engineering, Shenzhen University, Shenzhen, China}
\address[myfifthhaddress]{Guangdong Key Laboratory of Intelligent Information Processing, Shenzhen University, Shenzhen, China}
\address[mysecondaryaddress]{College of Computer Science and Software Engineering, Shenzhen University, Shenzhen, China}
\address[myfourthaddress]{Key Laboratory of Spatial Information Smarting Sensing and Services, Shenzhen University, Shenzhen, China}
\address[mythirdaddress]{School of Computer Science, University of Nottingham, Nottingham, United Kingdom}

\begin{abstract}
We present a new method for improving the performances of variational autoencoder (VAE). In addition to enforcing the deep feature consistent principle thus ensuring the VAE output and its corresponding input images to have similar deep features, we also implement a generative adversarial training mechanism to force the VAE to output realistic and natural images. We present experimental results to show that the VAE trained with our new method outperforms state of the art in generating face images with much clearer and more natural noses, eyes, teeth, hair textures as well as reasonable backgrounds. We also show that our method can learn powerful embeddings of input face images, which can be used to achieve facial attribute manipulation. Moreover we propose a multi-view feature extraction strategy to extract effective image representations, which can be used to achieve state of the art performance in facial attribute prediction.

\end{abstract}


\end{frontmatter}

\section{Introduction}

Deep convolutional neural networks (CNNs) \cite{GU2018354} have been used to achieve state of the art performances in many computer vision and image processing tasks such as image classification \cite{krizhevsky2012imagenet,simonyan2014very,he2016deep}, retrieval \cite{babenko2014neural}, detection \cite{girshick2014rich}, captioning \cite{karpathy2015deep}, human pose recovery \cite{yu2018multitask,hong2016hypergraph,hong2015image}, image privacy protection \cite{yu2017iprivacy}, unsupervised dimension reduction \cite{zhang2018local} and many other applications \cite{liu2018balance,junior2018randomized,osipov2018space}. Deep convolutional generative models, as a branch of unsupervised learning technique in machine learning, have become an area of active research in recent years. A generative model trained with a given image database can be useful in several ways. One is to learn the essence of a dataset and generate realistic images similar to those in the dataset from random inputs. The whole dataset is ``compressed"  into the learned parameters of the model, which are significantly smaller than the size of the training dataset. The other is to learn reusable feature representations from unlabeled image datasets for a variety of supervised learning tasks such as image classification.

In this paper, we propose a new method to train the variational autoencoder (VAE) \cite{kingma2013auto} to improve its performance. In particular, we seek to improve the quality of the generated images to make them more realistic and less blurry. To achieve this, we employ objective functions based on deep feature consistent principle \cite{hou2017deep} and generative adversarial network \cite{goodfellow2014generative,arjovsky2017wasserstein} instead of the problematic per-pixel loss functions. The deep feature consistent can help capture important perceptual features such as spatial correlation through the learned convolutional operations, while the adversarial training helps to produce images that reside on the manifold of natural images. We also introduce several techniques to improve the convergence of GAN training in this context. In particular, instead of directly using the generated images and the real images in pixel space, the corresponding deep features extracted from pretrained networks are used to train the generator and the discriminator network. We also propose to further relax the constraint on the output of the discriminator network to balance the image reconstruction loss and the adversarial loss. We present experimental results to show that our new method can generate face images with much clearer facial parts such as eyes, nose, mouth, teeth, ears and hair textures. We show that the VAE trained by our method can capture the semantic information of facial attributes, which can be modeled linearly in the learned latent space. Furthermore, we show that the trained VAE can be used to extract more discriminative facial attribute representations that can be used to achieve state of the art performance in facial attribute recognition. Concretely, our contributions are threefold: 
\begin{itemize} 
\item Our model seamlessly associates the two modalities, i.e., VAE and GAN through a common latent embedding space and we validate the effectiveness of this approach on image generation tasks.
\item We show that the learned latent representations can capture conceptual and semantic information of the input face images, which can be used to achieve facial attribute manipulation.
\item Lastly we introduce a multi-view feature extraction strategy on facial attribute recognition experiments in which we surpass state of the art.
\end{itemize}

The rest of the paper is organized as follows. We first briefly review the related literature in Section \ref{sec:related_work}. Section \ref{sec:method} presents our method to improve variational autoencoder with deep feature consistent and generative adversarial training. Section \ref{sec:experiments} presents experimental results which show that our method stands out as a state of the art technique. Finally we present a discussion and conclude the paper in Section \ref{sec:disscussion} and Section \ref{sec:conclusion}.


\section{Related Work}
\label{sec:related_work}
\subsection{Variational autoencoder}
Deep convolutional autoencoder is a powerful learning model for representation learning and has been widely used for different applications \cite{yu2018multitask,makkie2019fast,chen2018cross,chen2018evolutionary,feng2018graph,sun2017generalized,hong2015multimodal,hong2016hypergraph}. Variational Autoencoder (VAE) \cite{kingma2013auto,rezende2014stochastic} has become a popular generative model, allowing us to formalize image generation task in the framework of probabilistic graphical models with latent variables. Firstly it encodes an input image $x$ to a latent vector $z = E(x) \sim q(z|x)$ with an encoder network $E$, and then a decoder network $D$ is used to decode the latent vector $z$ back to image space, i.e., $\bar{x} = D(z) \sim p(x|z)$. In order to achieve image reconstruction we need to maximize the marginal log-likelihood of each observation (pixel) in $x$, and the VAE reconstruction loss $\mathcal{L}_{rec}$ is the negative expected log-likelihood of the observations in $x$. Another key property of VAE is the ability to control the distribution of the latent vector $z$, which has characteristic of being independent unit Gaussian random variable, i.e., $z \sim \mathcal{N}(0, I)$. Moreover, the difference between the distribution of $q(z|x)$ and the distribution of a Gaussian distribution (called KL Divergence) can be quantified and minimized by gradient descent algorithm \cite{kingma2013auto}. Therefore, VAE models can be trained by optimizing both of the reconstruction loss $\mathcal{L}_{rec}$ and KL divergence loss $\mathcal{L}_{kl}$.

\begin{equation}
\mathcal{L}_{rec} = - \mathbb{E}_{q(z|x)} [\log p(x|z)]
\end{equation}

\begin{equation}
\mathcal{L}_{kl} = D_{kl}(q(z|x)||p(z))
\end{equation}

\begin{equation}
\mathcal{L}_{vae} = \mathcal{L}_{rec} + \mathcal{L}_{kl}
\end{equation}

Several methods have been proposed to improve the performance of VAE. \cite{kingma2014semi} and \cite{yan2015attribute2image} proposed to build variational autoencoders by conditioning on either class labels or on a variety of visual attributes, and their experiments demonstrate that they are capable of generating realistic faces with diverse appearances.  Deep Recurrent Attentive Writer (DRAW) \cite{gregor2015draw} combines spatial attention mechanism with a sequential variational auto-encoding framework that allows iterative generation of images. \cite{ridgeway2015learning} and \cite{hou2017deep} consider replacing per-pixel loss with perceptual similarities using either multi-scale structural similarity score or a perceptual loss based on deep features extracted from pretrained deep networks.

\subsection{Generative adversarial network}

Generative Adversarial Network (GAN) framework is firstly introduced by \cite{goodfellow2014generative} to estimate generative models based on a min-max game. Under the GAN framework two models are simultaneously trained: a generator network $G(z)$ used to map a noise variable $z$ to data space, a discriminator network $Dis(x)$ designed to distinguish between the samples from the true training data and generated samples produced by the generator $G(z)$. The discriminator $Dis(x)$ is optimized by maximizing the probability of assigning the correct label for each category. The generator network $G(z)$ is trained simultaneously to minimize $log(1 - Dis(G(z)))$ by playing against the adversarial discriminator network $Dis(x)$. Thus the min-max game between $G(z)$ and $Dis(x)$ can be formulated as follows:
\begin{equation}
\min_G \max_{Dis} V(Dis, G) = \mathbb{E}_{x}[log(Dis(x))] + \mathbb{E}_{z}[log(1 - Dis(G(z)))]
\end{equation}
Following works \cite{arjovsky2017wasserstein,denton2015deep,radford2015unsupervised,im2016generating,salimans2016improved,chen2016infogan,karras2018progressive,hu2018unifying,dong2018san,DBLP:conf/cvpr/DongHYY17,DBLP:conf/cvpr/WanPGY17,DBLP:journals/corr/Qi17} have focused on improving the perceptual quality of GAN outputs and the training stability of GAN through architectural innovations and new training techniques. Our model enjoys both the advantages of deep feature consistent VAE (DFC-VAE) \cite{hou2017deep} and Wasserstein GAN (WGAN) \cite{arjovsky2017wasserstein} to improve the perceptual quality of the output images generated by VAE and enhance the effectiveness of VAE representations for semi-supervised learning. In addition, a combination of VAE and GAN was also proposed by \cite{larsen2015autoencoding}. Whilst there is a similarity, there are some differences as well. We use a pre-trained VGGNet as feature extractor to extract features of the input image and the output image and calculate the loss function. In reference \cite{larsen2015autoencoding}, they used the GAN discriminator network to extract image features to calculate the loss function and this discriminator was updated during the GAN training. Additionally, we adopt the framework of WGAN \cite{arjovsky2017wasserstein} to achieve adversarial training while  DCGAN \cite{radford2015unsupervised} was adopted in \cite{larsen2015autoencoding}.

\subsection{Learned features for image synthesis}
Neural style transfer \cite{gatys2015neural} is among the most successful applications of image synthesis based on the learned convolutional features in recent years. It tries to combine the content of one image with the style of another image by jointly optimizing content reconstruction loss and style reconstruction loss based on the features extracted from a pretrained convolutional neural network. Other works try to train a feed-forward network for real-time style transfer \cite{johnson2016perceptual,ulyanov2016texture,li2016combining}. In addition, images can be also generated by maximizing classification scores or individual features \cite{simonyan2013deep,yosinski2015understanding} for a better understanding of the trained networks. Furthermore high-confidence fooling images can be also synthesized through a similar optimizing technique \cite{szegedy2013intriguing,nguyen2015deep}.

\begin{figure}[!tb]
\centering
\begin{tabular}{ccc}
\includegraphics[width=12cm]{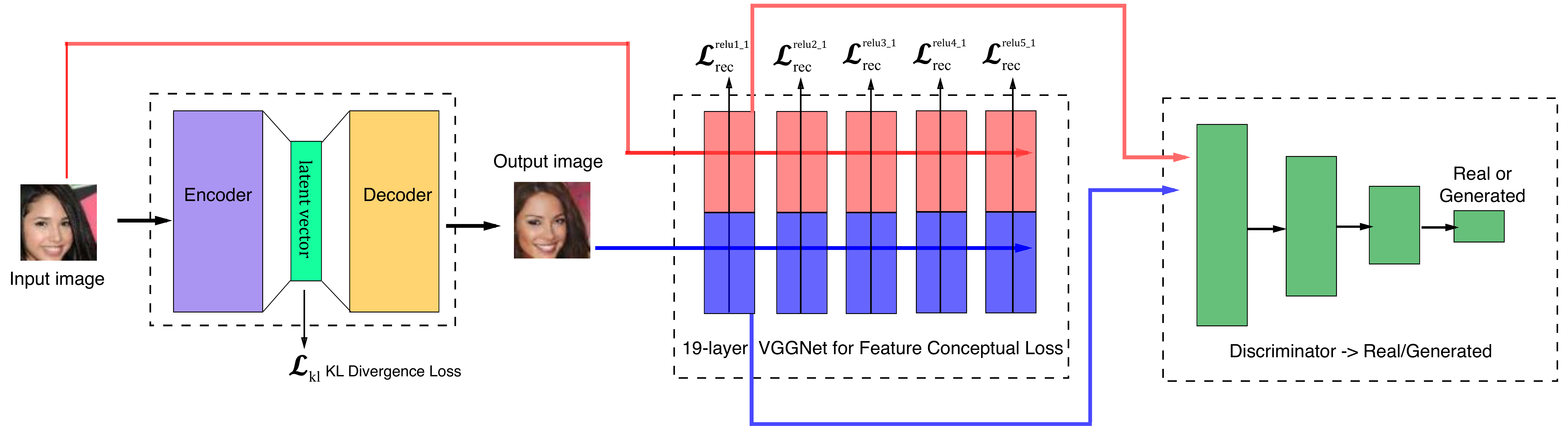}
\end{tabular}
\caption{Model overview. From left to right: The Variational autoencoder (VAE), the VGGNet used for feature extraction and WGAN discriminator. Note that the inputs fed to the discriminator come from the first convolutional layer of the VGGNet.}
\label{fig:overview}
\end{figure}

\section{Method}
\label{sec:method}
\subsection{Overview}
As shown in Figure \ref{fig:overview}, our model consists of three components: a variational autoencoder including an encoder network $E(x)$ and a decoder network $D(z)$, a pretrained VGGNet $\Phi(x)$ for feature extraction and a classifier network used as discriminator $Dis(x)$. Both the encoder and the decoder are deep residual convolutional neural networks with a 100-dimensional latent vector. The encoder processes the input image into the latent vector which is then decoded to an output image. In order to train a VAE, we need two losses, one is KL divergence loss $\mathcal{L}_{kl} = D_{kl}(q(z|x)||p(z))$ \cite{kingma2013auto}, which is used to make sure that the latent vector $z$ is an independent unit Gaussian random variable. The other is a feature reconstruction loss, which is based on the features extracted from VGGNet. Specifically we feed both of the input and output images to the pre-trained network $\Phi$ respectively and then measure the difference between the hidden layer representations, i.e., $\mathcal{L}_{rec} = \mathcal{L}^1 + \mathcal{L}^2 + ... + \mathcal{L}^l$, where $\mathcal{L}^l$ represents the feature reconstruction loss at the $l^{th}$ hidden layer. Furthermore, the VAE also serves as the generator and works with the discriminator to play the GAN game. Instead of feeding the pixels to the discriminator, we propose to use the first layer's output of the VGGNet as the input of the discriminator. The purpose is to enable more stable training as well as use as much low level image information as possible. It is worth noting that our architecture is different from that of \cite{larsen2015autoencoding}. Whilst they use the hidden layer features of the GAN discriminator to compute the image reconstruction loss, we adopt a pre-trained VGGNet. What's more, the pre-trained VGGNet is fixed during training and it still allows feed-forward and back-propagation computation. As a result, our model can be trained end-to-end.

\begin{figure}[!tb]
\begin{tabular}{ccc}
\includegraphics[width=12cm]{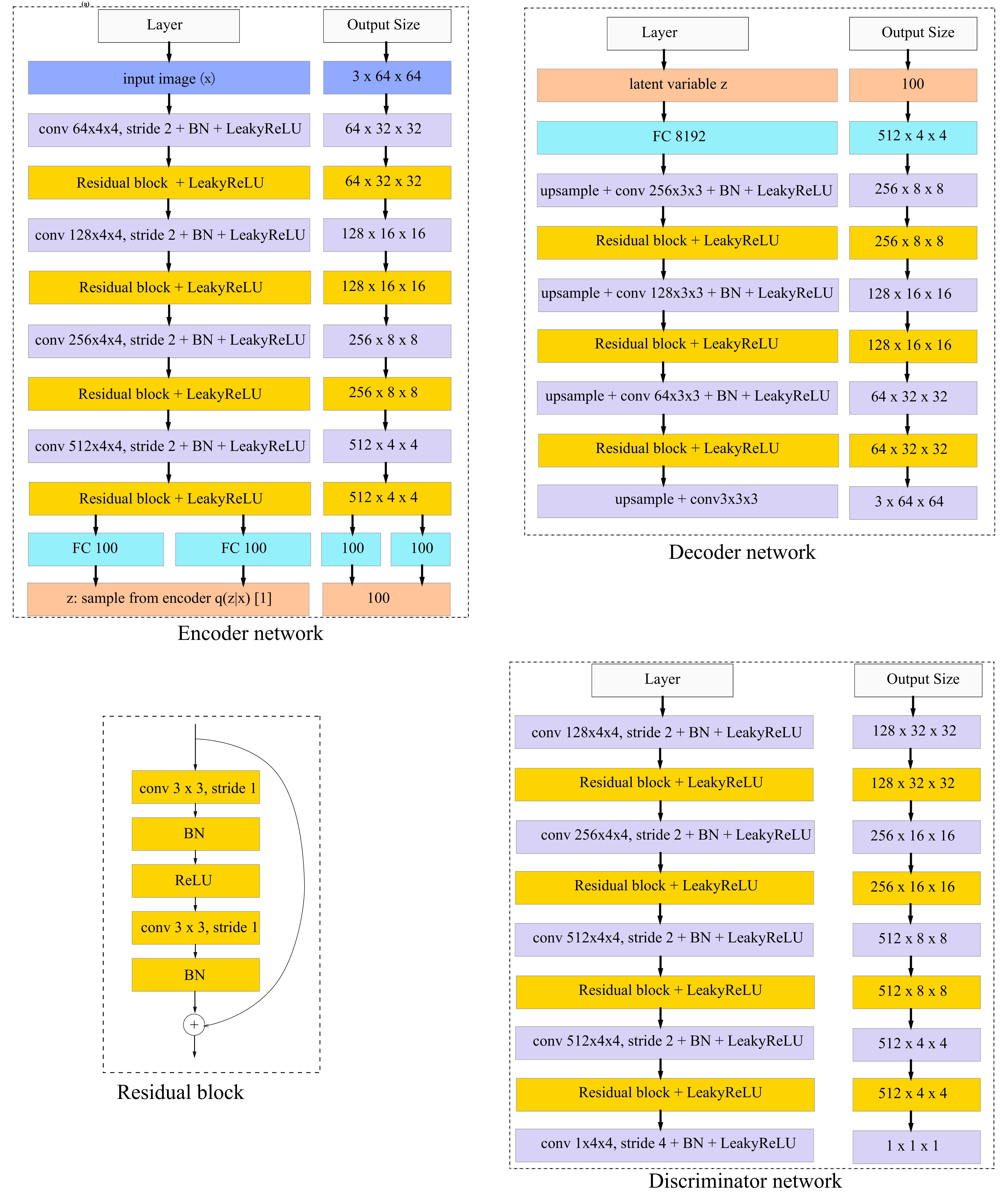}
\end{tabular}
\caption{The architecture of the autoencoder and discriminator network.}
\label{fig:autoencoder}
\end{figure}

\subsection{Neural network architecture}
As shown in Figure \ref{fig:autoencoder}, both of the autoencoder and discriminator network are deep residual convolutional neural networks based on \cite{he2016deep,radford2015unsupervised}. We construct 4 convolutional layers in the encoder network with 4 $\times$ 4 kernel and 2 $\times$ 2 stride to achieve spatial downsampling instead of using deterministic spatial functions such as maxpooling. Each convolutional layer is followed by a batch normalization layer and a LeakyReLU activation layer. In addition, a residual block is added after each convolutional layer and all the residual blocks contain two 3 $\times$ 3 kernel convolutional layers with the same number of filters. Lastly two fully-connected output layers (for mean and variance) are added to the encoder and will be used to calculate the KL divergence loss and sample latent variable $z$ (see \cite{kingma2013auto} for details).

For the decoder, we use 4 convolutional layers with 3 $\times$ 3 kernels and 1 $\times$ 1 stride. We also propose to replace standard zero-padding with replication padding, i.e., feature map of an input is padded with the replication of the input boundary. Similar to the encoder, each convolutional layer is also followed by a residual layer except the last one. For upsampling we use nearest neighbor method by a scale of 2 instead of fractional-strided convolutions used by other works \cite{long2015fully,radford2015unsupervised}. We also use batch normalization to help stabilize the whole training and use LeakyReLU as the activation function.

The design of the discriminator follows the architectural innovations of DCGAN \cite{radford2015unsupervised}. We use convolutional layers with 4 $\times$ 4 kernel and 2 $\times$ 2 stride to achieve spatial downsampling and add a residual block after each convolutional layer except the last layer. Like WGAN \cite{arjovsky2017wasserstein}, the sigmoid layer is removed in the last layer and use a 4 $\times$ 4 stride convolution layer to produce a single output, and the gradients of discriminator is clipped between -0.01 to 0.01.

\subsection{Feature reconstruction loss}
Feature reconstruction loss of two images is defined as the difference between the hidden features in a pretrained deep convolutional neural network $\Phi$. Similar to \cite{gatys2015neural}, we use VGGNet \cite{simonyan2014very} as the loss network in our experiment. The core idea of feature reconstruction loss is to seek consistency between two images in the learned feature space. As the hidden representations can capture important perceptual quality features such as spatial correlation, a smaller difference of hidden representations indicates a better consistency of spatial correlations between the input and the output, as a result, we can get a better visual quality of the output image.
Specifically, let $\Phi_l(x)$ denotes the representation of the $l^{th}$ hidden layer when input image $x$ is fed to network $\Phi$. Mathematically $\Phi_l(x)$ is a 3D volume block array of shape [$C_l$ x $W_l$ x $H_l$], where $C_l$ is the number of filters, $W_l$ and $H_l$ denote the width and height of each feature map for the $l^{th}$ layer. The feature reconstruction loss for one layer ($\mathcal{L}_l$) between two images $x$ and $\bar{x}$ can be simply defined by squared Euclidean distance. Actually it is quite like the per-pixel loss for images except that the number of color channels is not 3 anymore.

\begin{equation}
\mathcal{L}_l =  \frac{1}{2C_lW_lH_l}  \sum_{c=1}^{C_l}  \sum_{w=1}^{W_l}  \sum_{h=1}^{H_l} (\Phi_l(x)_{c,w,h} - \Phi_l(\bar{x})_{c,w,h})^2
\end{equation}

Instead of only using a single layer features, we leverage visual features in different layers and combine the outputs of the five convolutional layers of the VGGNet. The final reconstruction loss is defined as:
\begin{equation}
\mathcal{L}_{rec} = \sum_{l=1}^L \frac{100} {C_l^2} \mathcal{L}_l
\end{equation}
where $\mathcal{L}_l$ and $C_l$ are the feature loss and the number of filters at $l^{th}$ layer respectively, $L$ is total convolutional layers in the pretrained network.

Additionally we adopt the KL divergence loss $\mathcal{L}_{kl}$ \cite{kingma2013auto} to regularize the encoder network to control the distribution of the latent variable $z$. To train VAE, we jointly minimize the KL divergence loss $\mathcal{L}_{kl}$ and the feature reconstruction loss $\mathcal{L}_{rec}$ for different layers as follows:

\begin{equation}
\label{eq:vae_loss}
\mathcal{L}_{vae} = \alpha \mathcal{L}_{kl} + \beta \mathcal{L}_{rec}
\end{equation}

where $\alpha$ and $\beta$ are the weighting parameters for KL Divergence loss and feature reconstruction loss. It is worth noting that the pre-trained VGGNet is used for feature extraction only and is fixed during the training. The latent representation of the image refers to the latent variable of the autoencoder in our paper.

\subsection{Adversarial loss}
In addition to the feature reconstruction loss described above, we also incorporate variational autoencoder in the framework of generative adversarial network to encourage the VAE to produce outputs that reside on the manifold of natural images. Our adversarial training is based on WGAN \cite{arjovsky2017wasserstein}. In order to further improve the training stability, instead of directly feeding the real images and generated images to a discriminator, we first extract the first layer features of the pretrained VGGNet and feed them to the discriminator network. It is because we would like to push the reconstructed image similar to natural images in terms of low-level information, which can be often obtained from lower layers of deep networks. In addition, we propose another technique to further relax the constraint on the output of the discriminator network. WGAN \cite{arjovsky2017wasserstein} proposes to remove the last Sigmoid layer in the generator and use 1 and -1 as ground-truth label for real and generated images. In our experiments, we found that GAN training could collapse and the VAE training tends to dominate the training when using too small labels, e.g., 1 and -1. In addition, we also found that the adversarial loss would dominate the training by using too big labels like -100 and 100, which could lead to structural changes of the reconstructed images. Using empirical values 10 and -10 to represent ground-truth labels, we can effectively balance well between the VAE and GAN, and generate diverse synthesized results in a more natural and flexible manner.

Finally our entire deep model can be trained end-to-end with a combination of KL divergence loss, reconstruction loss and adversarial loss as Equation \ref{eq:loss} and the training procedure is summarized in Algorithm 1.

\begin{equation}
\label{eq:loss}
\mathcal{L}_{vae} = \alpha \mathcal{L}_{kl} + \beta \mathcal{L}_{rec} + \mathcal{L}_{GAN}
\end{equation}

\begin{algorithm}
\caption{Training VAE-WGAN Model}
\begin{algorithmic} 
\REQUIRE \text{c, the clipping parameter}; \text{$\Phi$, pretrained model}
\STATE 	$\textit{W}_{Encoder}, \textit{W}_{Decoder}, \textit{W}_{Discriminator} \leftarrow \text{ Initialize parameters }$
\REPEAT
\STATE 	$X \leftarrow \text{random mini-batch images from the dataset}$
\STATE  $Z \leftarrow \text{Encoder($X$)}$
\STATE  $\mathcal{L}_{kl} \leftarrow D_{KL} \left( q(Z|X) || p(Z)\right)$
\STATE  $\hat{X} \leftarrow \text{Decoder($Z$)}$
\STATE  $\mathcal{L}_{rec} \leftarrow ||\Phi(X) - \Phi(\hat{X})||^2$
\STATE  $\mathcal{L}_{GAN} \leftarrow Discriminator(X) - Discriminator(\hat{X})$ // Wasserstein GAN

\STATE  $\textit{W}_{Encoder} \stackrel{+}\leftarrow - \nabla_{\textit{W}_\text{Encoder}} \left(\mathcal{L}_{kl} + \mathcal{L}_{rec} - \mathcal{L}_{GAN} \right) $
\STATE  $\textit{W}_{Decoder} \stackrel{+}\leftarrow - \nabla_{\textit{W}_\text{Decoder}} \left(\mathcal{L}_{rec} - \mathcal{L}_{GAN} \right) $

\STATE  $\textit{W}_{Discriminator} \stackrel{+}\leftarrow - \nabla_{\textit{W}_\text{Discriminator} \: \mathcal{L}_{GAN}}$
\STATE  $\textit{W}_{Discriminator} \leftarrow clip\left(\textit{W}_{Discriminator}, -c, c \right)$

\UNTIL \text {convergence of parameters}

\end{algorithmic}
\end{algorithm}

\section{Experiments}
\label{sec:experiments}
In this paper, we conduct experiments on CelebFaces Attributes (CelebA) \cite{liu2015deep} and CIFAR-10 \cite{Krizhevsky09} Dataset to evaluate our method on the performance of image generation. We also study how different layer features of the pre-trained VGGNet affects the performances of image synthesis. Furthermore, we consider manipulating the facial attributes in the learned latent space. Finally we apply the learned representations to facial attribute recognition and show that we can achieve state of the art performances.

\subsection{Training details}
CelebA is a large-scale face attribute dataset with 202,599 face images, 5 landmark locations and 40 binary attributes annotations per image. We build the training dataset by cropping and scaling the aligned images to 64 $\times$ 64 pixels like \cite{larsen2015autoencoding,radford2015unsupervised}. The CIFAR-10 dataset consists of 60,000 images of shape 32 $\times$ 32 in 10 classes. There are 50,000 training images and 10,000 test images. For both datasets, we train our model with a batch size of 64 for 5 epochs over the training dataset and use Adam method for optimization \cite{kingma2014adam} with an initial learning rate of 0.0005, which is decreased by a factor of 0.5 for the following epochs. The 19-layer VGGNet \cite{simonyan2014very} is chosen as loss network $\Phi$ to construct feature reconstruction loss for image reconstruction. The loss weighting parameters $\alpha$ and $\beta$ are 1 and 0.5 respectively. Our implementation is built on deep learning framework Torch \cite{collobert2011torch7}. As for the computational time, it takes around 10 hours to train our models and 0.012 seconds to process an image of size 64 $\times$ 64 during testing. The training and testing time are both benchmarked on a single GTX 1080Ti GPU.

\subsection{Qualitative results for image generation}
The comparison is divided into two parts: one is arbitrary image generation decoded from vectors $z$ randomly drawn from $\mathcal{N}(0, 1)$, the other is natural image reconstruction.

\begin{figure}[!tb]
\begin{tabular}{ccc}
\centering
\includegraphics[width=11cm]{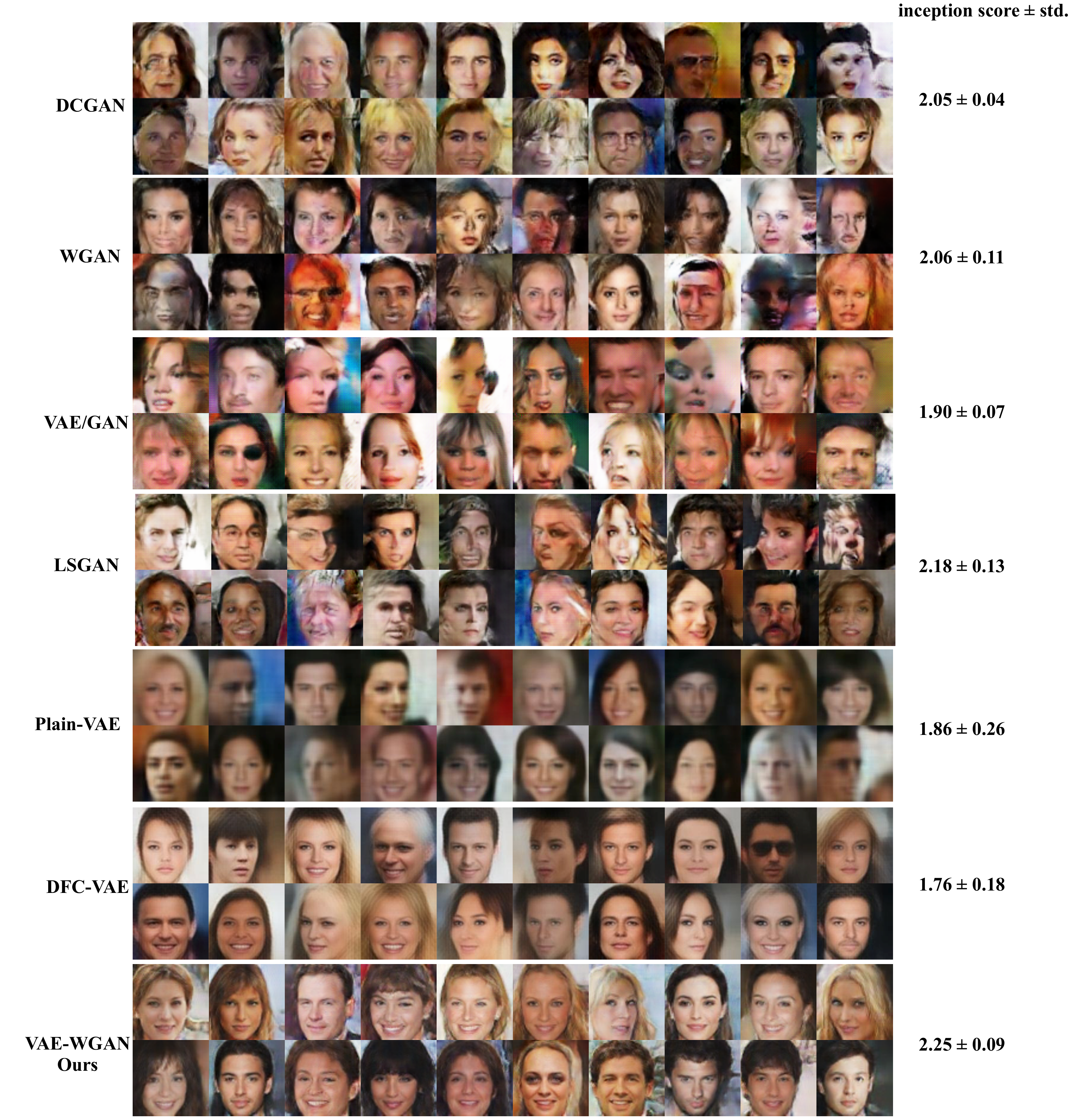}
\end{tabular}
\caption{Face images generated from 100-dimension latent vector $z \sim \mathcal{N}(0, 1)$ by different models. We compare our VAE-WGAN with DCGAN\cite{radford2015unsupervised}, WGAN\cite{arjovsky2017wasserstein}, VAE/GAN\cite{larsen2015autoencoding}, LSGAN\cite{DBLP:journals/corr/Qi17}, Plain-VAE\cite{kingma2013auto} and DFC-VAE\cite{hou2017deep}.}
\label{fig:random_faces}
\end{figure}

\begin{figure}[!tb]
\begin{tabular}{ccc}
\centering
\includegraphics[width=11cm]{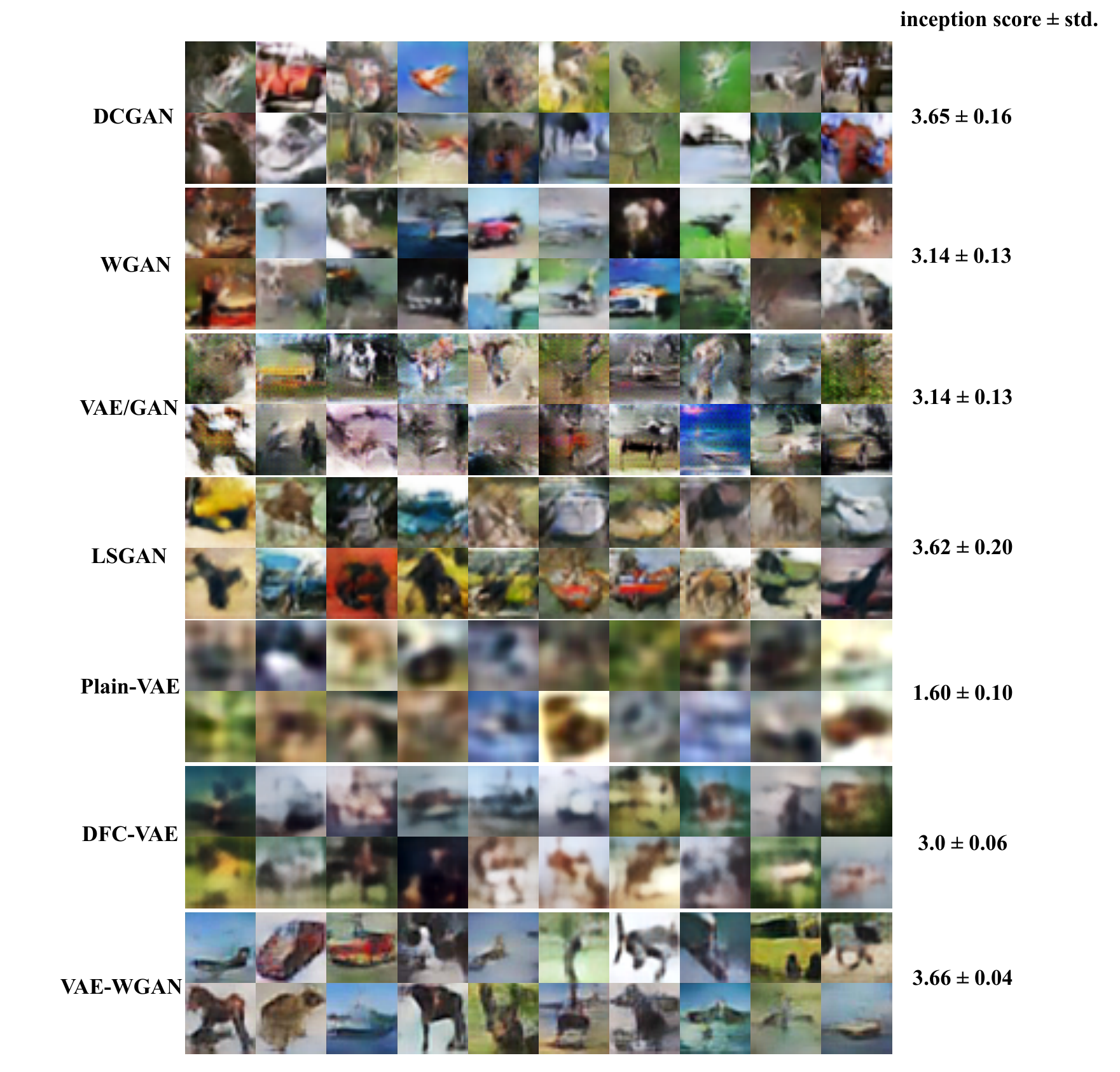}
\end{tabular}
\caption{Cifar images generated from 100-dimension latent vector $z \sim \mathcal{N}(0, 1)$ by different models. We compare our VAE-WGAN with DCGAN\cite{radford2015unsupervised}, WGAN\cite{arjovsky2017wasserstein}, VAE/GAN\cite{larsen2015autoencoding}, LSGAN\cite{DBLP:journals/corr/Qi17}, Plain-VAE\cite{kingma2013auto} and DFC-VAE\cite{hou2017deep}.}
\label{fig:random_cifar}
\end{figure}

\begin{figure}[!tb]
\begin{tabular}{ccc}
\centering
\includegraphics[width=11cm]{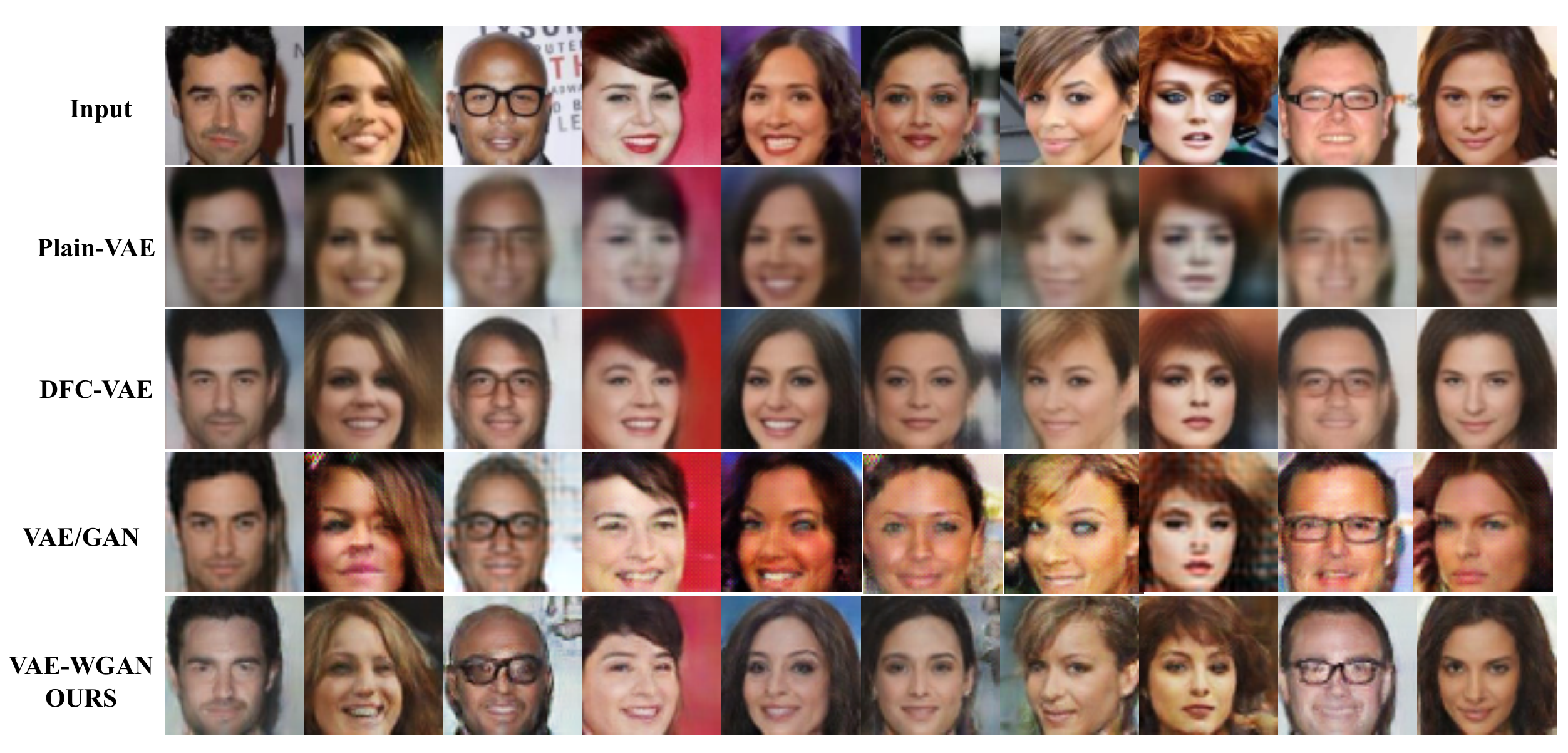}
\end{tabular}
\caption{Face images reconstructed by different models. We compare our VAE-WGAN with Plain-VAE\cite{kingma2013auto}, VAE/GAN\cite{larsen2015autoencoding} and DFC-VAE\cite{hou2017deep}.}
\label{fig:reconstruction}
\end{figure}

\begin{figure}[!tb]
\begin{tabular}{ccc}
\centering
\includegraphics[width=11cm]{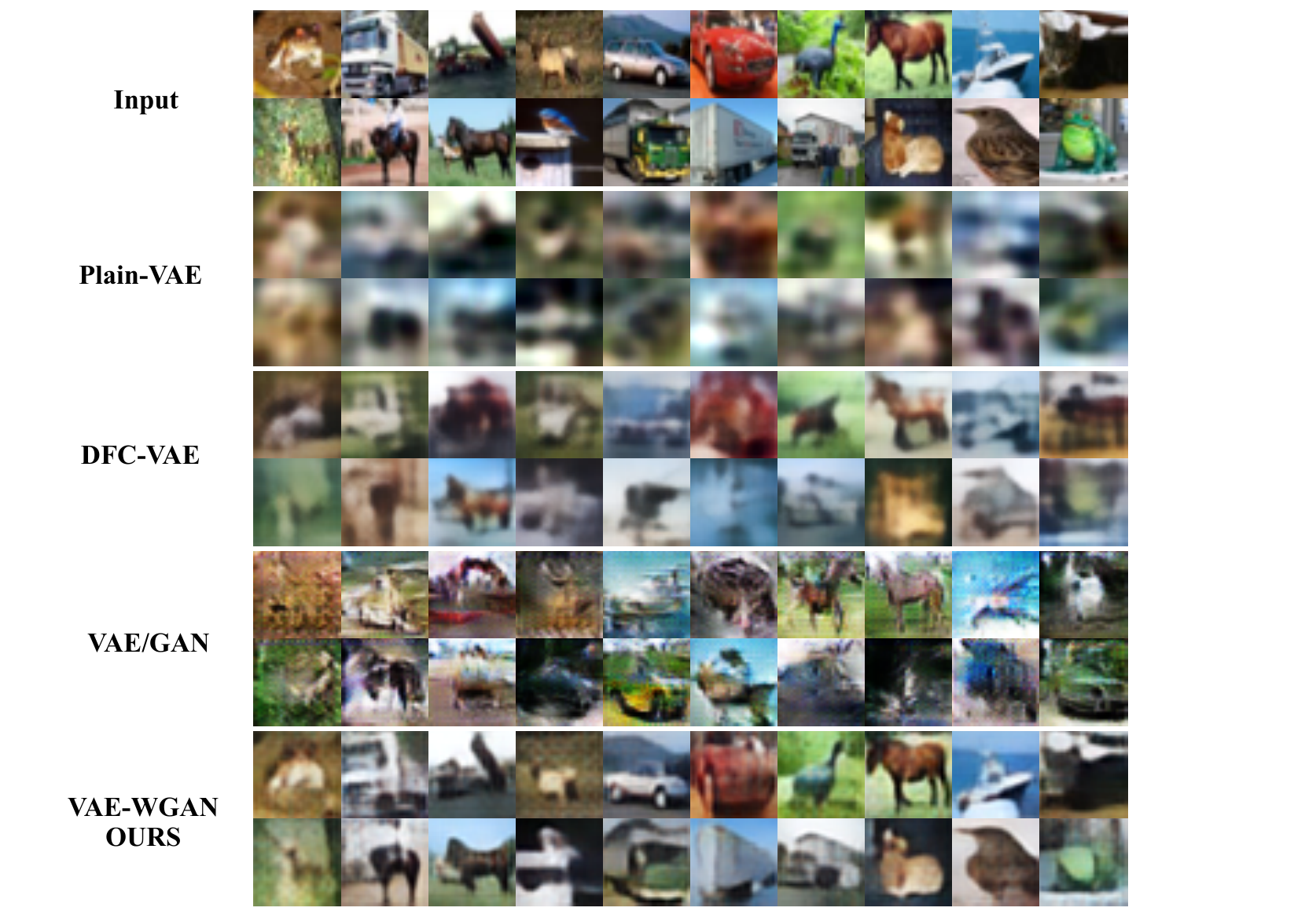}
\end{tabular}
\caption{CIFAR images reconstructed by different models. We compare our VAE-WGAN with Plain-VAE\cite{kingma2013auto}, VAE/GAN\cite{larsen2015autoencoding} and DFC-VAE\cite{hou2017deep}.}
\label{fig:reconstruction_cifar}
\end{figure}

\subsubsection{Arbitrary image generation.}

First, we compare the perceptual quality of the output face images for different generative models. As shown in Figure \ref{fig:random_faces} and \ref{fig:random_cifar}, we compare our model VAE-WGAN with Plain-VAE \cite{kingma2013auto}, DFC-VAE \cite{hou2017deep}, DCGAN \cite{radford2015unsupervised}, WGAN \cite{arjovsky2017wasserstein}, VAE/GAN \cite{larsen2015autoencoding} and LSGAN\cite{DBLP:journals/corr/Qi17}. All the compared models are implemented with the public available code from the corresponding papers with default settings. The final output images are produced by feeding vectors randomly drawn from a given distribution $\mathcal{N}(0, 1)$ to either VAE decoder or GAN generator. We can see that DCGAN, WGAN as well as LSGAN can generate clean and sharp images, however the image details can be distorted, resulting in unsatisfactory outputs with weird appearance like unpleasing faces. It is because there no input image information for pure GAN training. In contrast, the results produced by VAE decoder can better preserve the overall object structures. However, Plain-VAE tends to produce very blurry images because it tries to minimize the per-pixel loss between two images and each pixel is optimized independently. DFC-VAE can produce clear and sharp images because the feature reconstruction loss contains the perceptual and spatial correlation information in the learned feature space. VAE/GAN and our VAE-WGAN can achieve better results than all the other models, however VAE/GAN still suffers from observed distortions. Our method can generate more consistent and realistic human faces with much clearer noses, eyes, teeth, hair textures as well as reasonable backgrounds. Moreover, our method can achieve highest inception scores \cite{salimans2016improved} on the two dataset as shown in Figure \ref{fig:random_faces} and \ref{fig:random_cifar}. The inception scores are calculated based on 2,000 images for each model.

\subsubsection{Image reconstruction.}

We also evaluate the reconstruction performance of our method (shown in Figure \ref{fig:reconstruction} and \ref{fig:reconstruction_cifar}) by comparing with Plain-VAE, DFC-VAE \cite{hou2017deep} and VAE/GAN \cite{larsen2015autoencoding}. Pure GAN models are not involved because of no input images in their models. Similar to arbitrary images generated above, Plain-VAE reconstructs very blurry images because of the shortcomings of per-pixel loss. DFC-VAE can produce better images such as faces with clear eyes and mouths, however it still produces blurry background for CIFAR images and unrealistic hairs for face images. The results of VAE/GAN show that the images are reasonably sharp and clear, however details in the original images are missing. Again our model can produce much better reconstruction results than other models. Our model is better at preserving the original color and overall structures of the input images.

\begin{figure}[!tb]
\begin{tabular}{ccc}
\centering
\includegraphics[width=11cm]{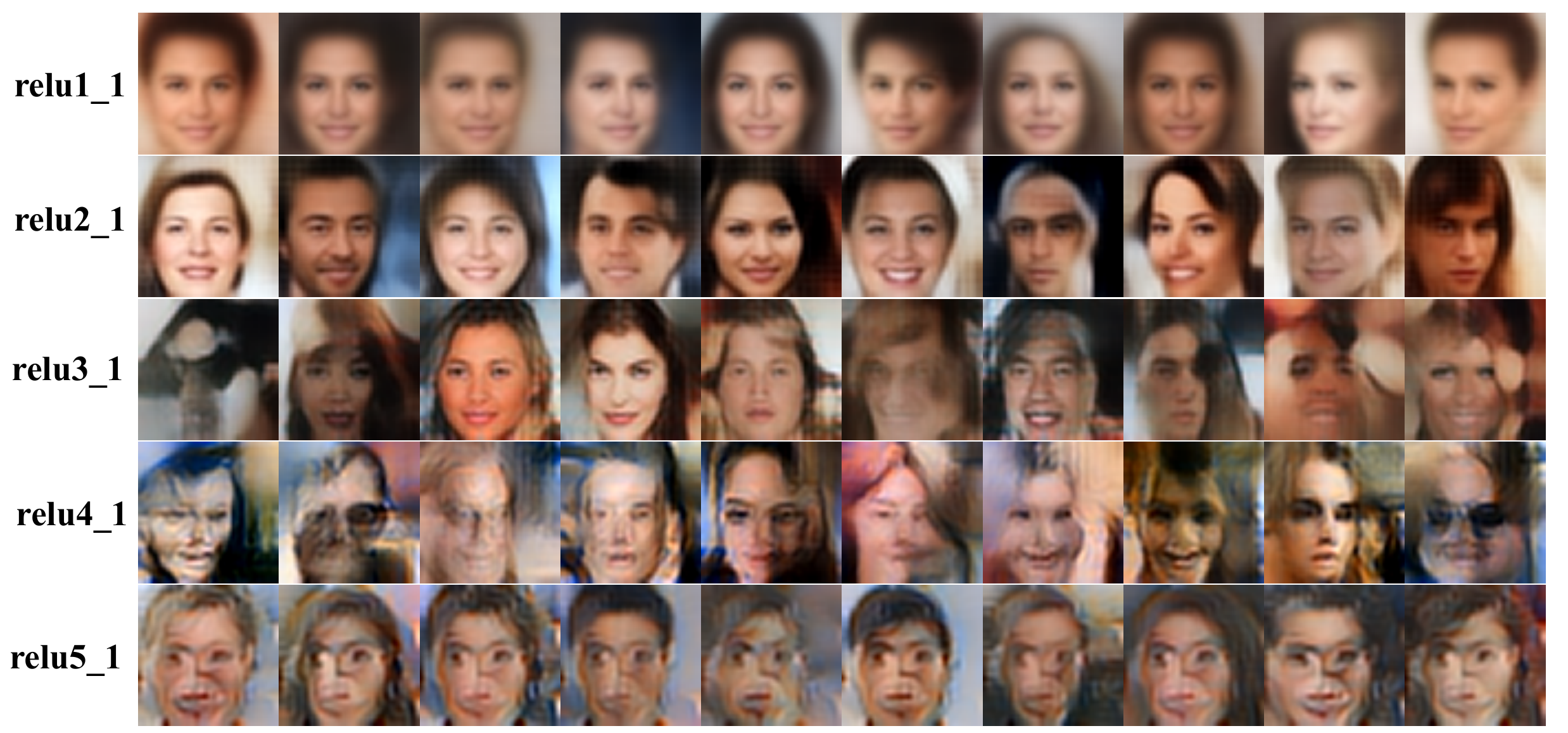}
\end{tabular}
\caption{Generated face images from 100-dimension latent vector $z \sim \mathcal{N}(0, 1)$ by 5 different models, which are trained with feature reconstruction loss based on layers relu1\_1, relu2\_1 relu3\_1, relu4\_1 and relu5\_1 respectively.}
\label{fig:random_face_multi_level}
\end{figure}

\subsubsection{Impact of different level reconstruction loss}
We also conduct experiments to investigate how features of different level convolutional layers of the loss network affect the quality of image generation. Figure \ref{fig:random_face_multi_level} shows the randomly generated face images by our five models trained with feature reconstruction loss based on layers relu1\_1, relu2\_1 relu3\_1, relu4\_1 and relu5\_1 respectively. It can be seen that all the generated images are able to keep the overall structures of faces. However as we reconstruct from lower level layers like relu1\_1, the generated images are very blurry especially in the hair and background area. When using higher level layers, the generated face images are much sharper and can show reasonable hair textures, but the exact structure of facial attributes cannot be preserved like eyes and mouths. One explanation for this is that the higher level features are corresponding to a coarser space area of the encoded image. The areas covered by relu4\_1 and conv5\_1 layers are too large to construct local facial attributes like mouth and eyes, but better for larger area textures like hair. Overall we can get better results when using reconstruction loss by combining different layers.

\subsubsection{Impact of weighting parameters $\alpha$ and $\beta$}
\begin{figure}[!tb]
\centering
\begin{tabular}{ccc}
\includegraphics[width=11cm]{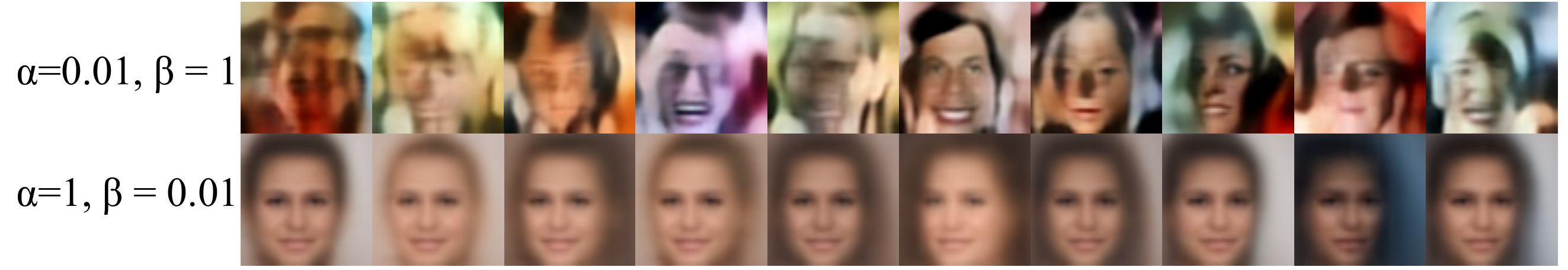}
\end{tabular}
\caption{Randomly generated images by our method with different weighting parameters $\alpha$ and $\beta$.}
\label{fig:random_neuro}
\end{figure}

\begin{figure}[!tb]
\centering
\begin{tabular}{ccc}
\includegraphics[width=11cm]{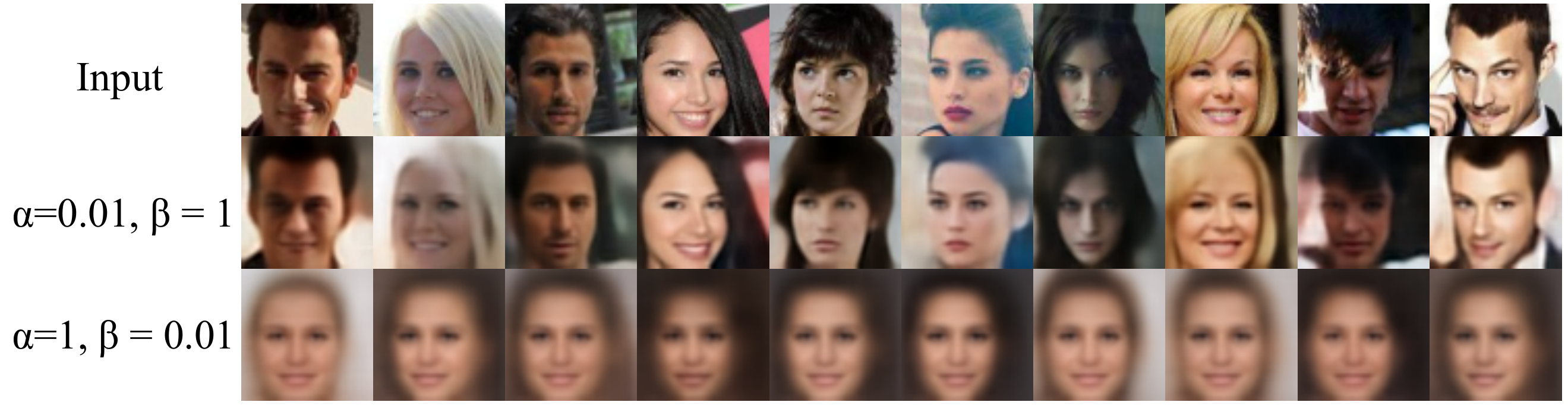}
\end{tabular}
\caption{Reconstruction results by our method with different weighting parameters $\alpha$ and $\beta$.}
\label{fig:reconstruction_neuro}
\end{figure}

We further conduct experiments to look into the influences of weighting parameters $\alpha$ and $\beta$ in Equation \ref{eq:vae_loss} in terms of image quality. Specifically we train two models with $\alpha = 1, \beta = 0.01$ and $\alpha = 0.01, \beta = 1$ respectively. As shown in Figure \ref{fig:random_neuro} and Figure \ref{fig:reconstruction_neuro}, we can see that the images can be better reconstructed when using bigger $\beta$, however the randomly generated images look weird with unusual face shapes. In addition, the randomly generated images are similar to the reconstructed ones with bigger $\alpha$ while they usually suffer from the problems of poor quality and lack of diversity. It is clear that the $\alpha$ and $\beta$ can be used to balance the trade-off  between the latent variable distribution and image reconstruction in variational autoencoder. As shown in previous sections, our model works well with $\alpha = 1$ and $\beta = 0.5$.

\begin{figure}[!tb]
\centering
\begin{tabular}{ccc}
\includegraphics[width=6cm]{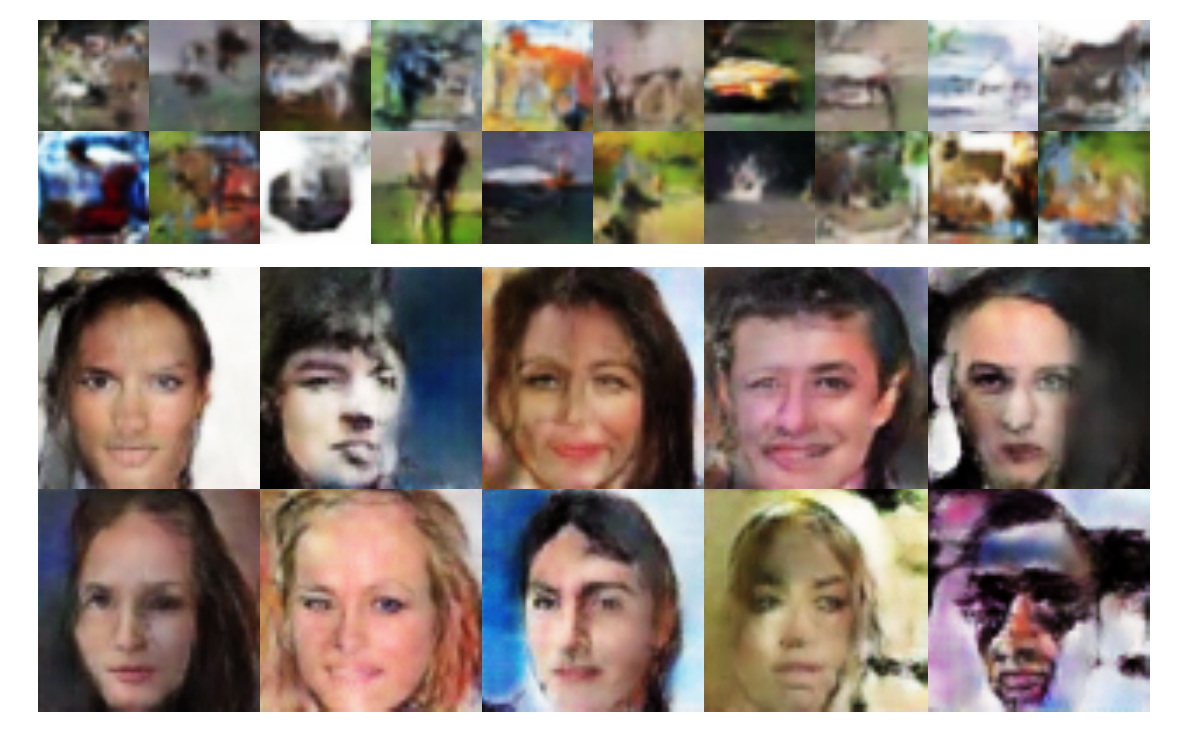}
\end{tabular}
\caption{Generated images by our method without reconstruction loss.}
\label{fig:random_face_no_reconstuction}
\end{figure}

In addition, we also conduct experiments without reconstruction loss. As shown in Figure \ref{fig:random_face_no_reconstuction}, we can see that the results are similar to those trained with DCGAN and WGAN. This is because the latent vector distribution is similar to the pre-defined Gaussian distribution without reconstruction constraint. Thus the whole training processing is roughly equal to a pure GAN training.

\subsection{The learned latent space}
In order to get a better understanding of what our model has learned, we investigate the property of the learned representation in the latent space. What's more, we also conduct experiments to show the effectiveness of our model to learn meaningful feature representations beyond image generation. In particular, we visualize the latent representations based on the t-SNE embedding and also apply them to the facial attribute recognition task.

\subsubsection{Linear interpolation of latent space}
\begin{figure}[!tb]
\centering
\includegraphics[width=11cm]{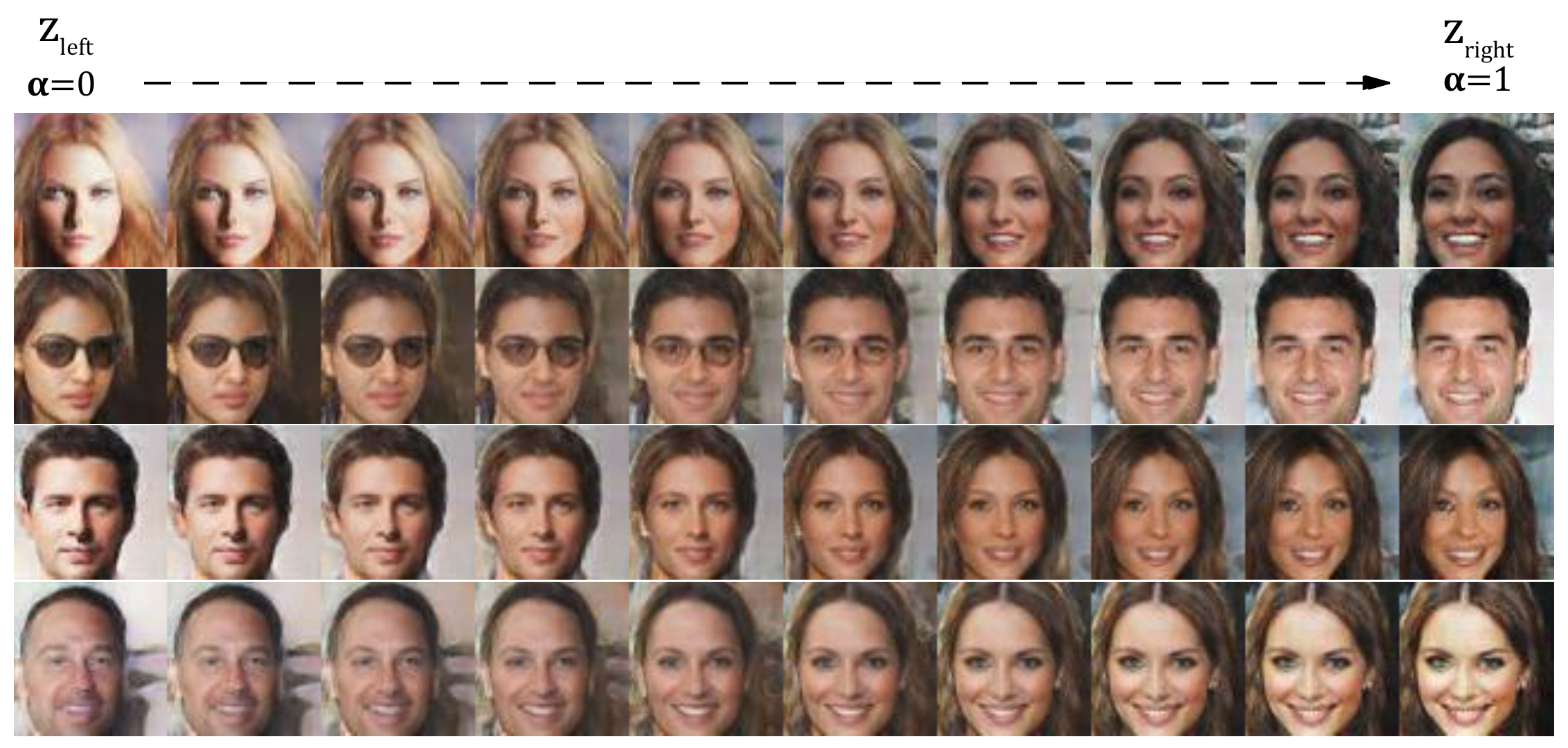}
\caption{Linear interpolation of latent vector. Each row is the interpolation from left latent vector $z_{left}$ to right latent vector $z_{right}$. e.g. $(1-\alpha) z_{left} + \alpha z_{right}$.}
\label{fig:linear_interpolation}
\end{figure}

As shown in Figure \ref{fig:linear_interpolation}, we have studied the linear interpolation between the generated images from two latent vectors denoted as $z_{left}$ and $z_{right}$. The interpolation is defined by a simple linear transformation $z = (1-\alpha) z_{left} + \alpha z_{right}$, where $\alpha = 0, 0.1, \dots, 1$, and then $z$ is fed to the decoder network to generate new face images. From the first row in Figure \ref{fig:linear_interpolation}, we can see the smooth transitions between $vector$(``Woman without smiling and blond hair") and $vector$(``Woman with smiling and black hair"). Little by little the color of the hair becomes black, the distance between lips becomes larger and teeth are shown in the end as smiling, and pose turns from looking slightly front to looking right. Additionally we provide examples of transitions between $vector$(``Man without eyeglass") and $vector$(``Woman with eyeglass"), as well as $vector$(``Man") and $vector$(``Woman").

\begin{figure}[!tb]
\centering
\begin{tabular}{ccc}
\includegraphics[width=11cm]{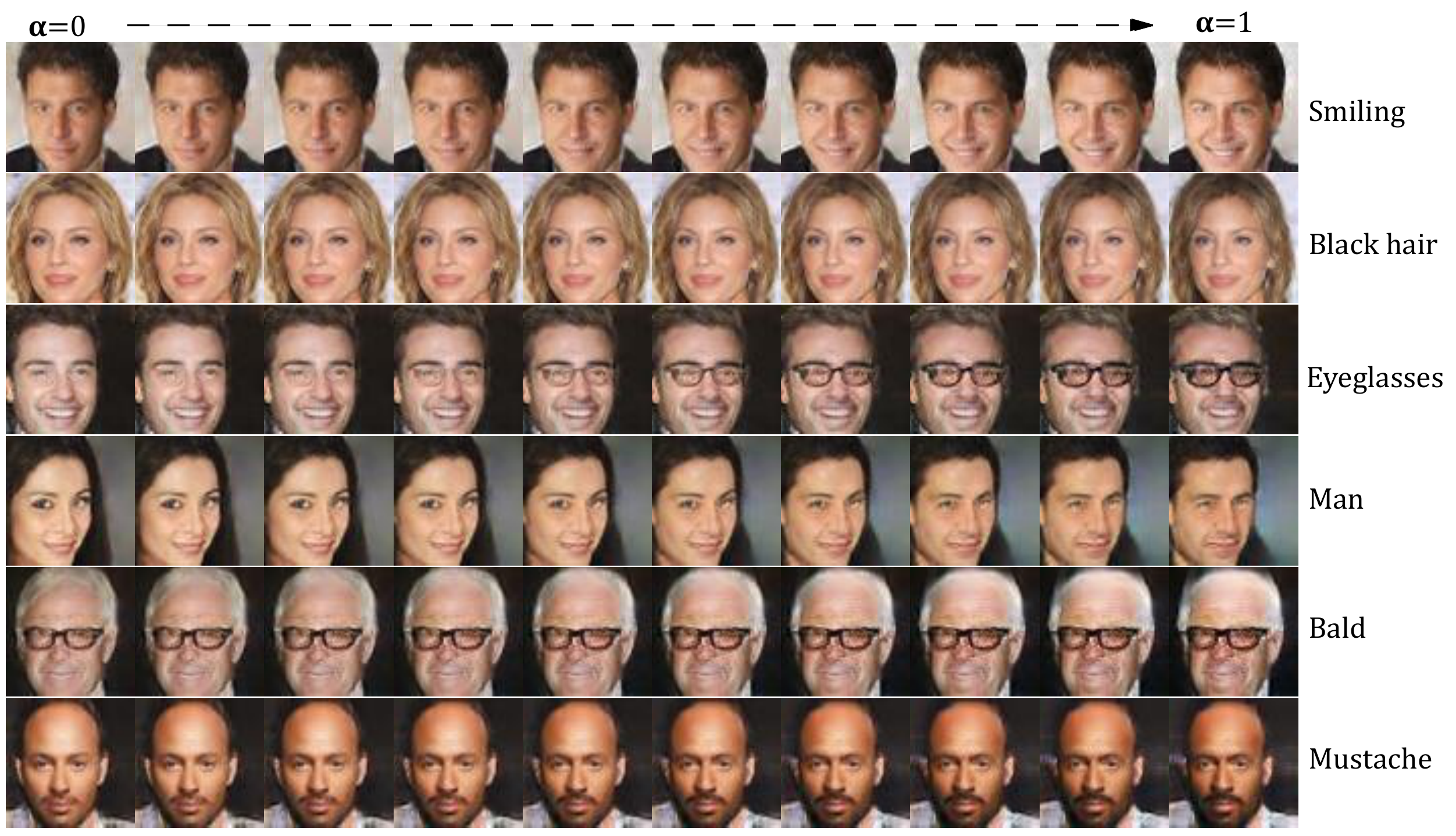}
\end{tabular}
\caption{Vector arithmetic for visual attributes. Each row is the generated faces from latent vector $z_{left}$ by adding or subtracting an attribute-specific vector, i.e., $z_{left}$ + $\alpha$ $z_{smiling}$, where $\alpha = 0, 0.1, \dots, 1$.}
\label{fig:attribute_specific}
\end{figure}


\subsubsection{Facial attribute manipulation}
The experiments above demonstrate interesting smooth transitional property between two latent vectors. In this section, instead of manipulating the overall face images, we seek to find a way to control a specific attribute of face images. In previous works, \cite{mikolov2013distributed} shows that $vector$(``King") - $vector$(``Man") + $vector$(``Woman") generates a vector whose nearest neighbor is the $vector$(``Queen") when evaluating learned representation of words. \cite{radford2015unsupervised} demonstrates that visual concepts such as face poses and gender could be manipulated by simple vector arithmetics.

In this paper, we conduct experiments to manipulate the facial attributes in the learned latent space of VAE-WGAN. For a given attribute such as \emph{smiling}, 2,000 smiling face samples are fed into the trained encoder to generate 2,000 latent vectors. The average of these vectors forms the latent representation $z_{smiling+}$. Similarly, we use 2,000 non-smiling face samples to generate a non-smiling latent vector $z_{smiling-}$. Finally the difference $z_{smiling} = z_{smiling+} - z_{smiling-}$, which in effect takes away any non-smiling attributes from the smiling images, is used as the semantic representation for the attribute \emph{smiling}. Similarly, we use the same approach to constructing other semantic attribute latent reconstructions for \emph{Bald}, \emph{Black hair}, \emph{Eyeglass}, \emph{Male} and \emph{Mustache}. Thus, for a given image with latent vector $z$, we can manipulate the facial attribute with the corresponding attribute vector arithmetically, e.g. $z = z + \alpha z_{smiling}$. Figure \ref{fig:attribute_specific} shows the results for the 6 attributes, i.e., \emph{Bald}, \emph{Black hair}, \emph{Eyeglass}, \emph{Male}, \emph{Smiling}, and \emph{Mustache}. As shown in Figure \ref{fig:attribute_specific}, by adding a smiling vector to the latent representation of a non-smiling man, we can observe the smooth transitions from non-smiling face to smiling face (the first row). Furthermore, the smiling appearance becomes more obvious when the weighting factor $\alpha$ is bigger, while other facial attributes are able to remain unchanged. We can see that our method can achieve smooth image transitions for different facial attributes with high quality, demonstrating that the face attributes can be modeled linearly in the learned latent space.


\begin{figure*}[!tb]
\centering
\begin{tabular}{ccc}
\includegraphics[width=12cm]{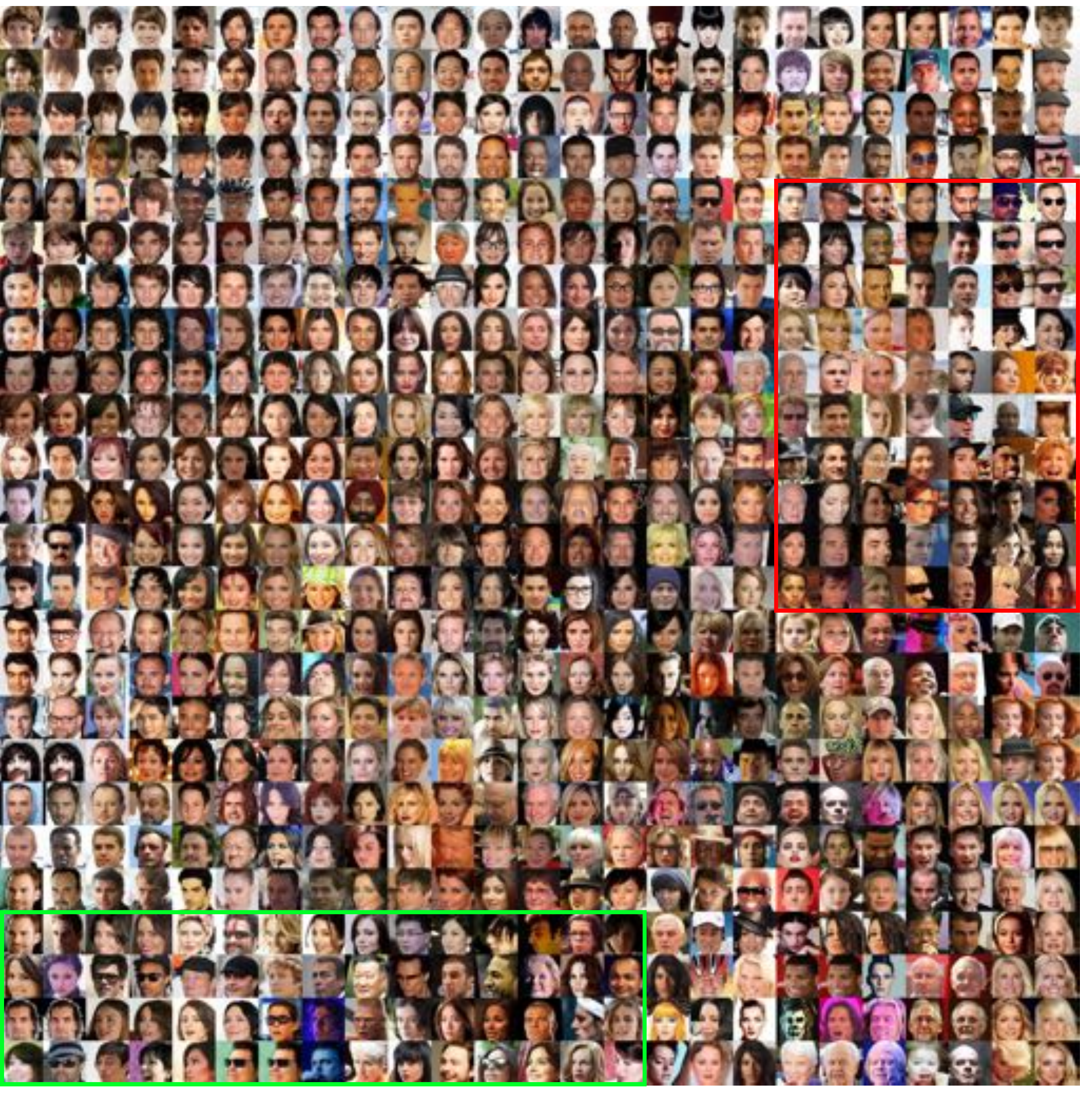}
\end{tabular}
\caption{Visualization of 25 x 25 face images based on latent vectors by t-SNE algorithm \cite{maaten2008visualizing}. }
\label{fig:cnn_embed_full_2k}
\end{figure*}


\subsubsection{Visualization of latent vectors}
Considering that the latent vectors are nothing but the encoding representation of the natural face images, it would be interesting to visualize the natural face images based on the similarity of their latent representations. Specifically we randomly choose 625 face images from CelebA dataset and extract the corresponding 100-dimensional latent vectors, which are then reduced to 2-dimensional embedding by t-SNE algorithm \cite{maaten2008visualizing}. t-SNE can arrange images that have similar high-dimensional vectors ($L_2$ distance) to be nearby each other in the embedding space. The visualization of 25 $\times$ 25 images is shown in Figure \ref{fig:cnn_embed_full_2k}. We can see that images with a similar background (black or white) tend to be clustered together. Furthermore, the face pose information can be also captured even no pose annotations in the dataset. The face images in the upper right (red rectangle) are those looking to the left and samples in the bottom left (green rectangle) are those looking to the right.

\begin{figure*}[!tb]
\centering
\begin{tabular}{ccc}
\includegraphics[width=11cm]{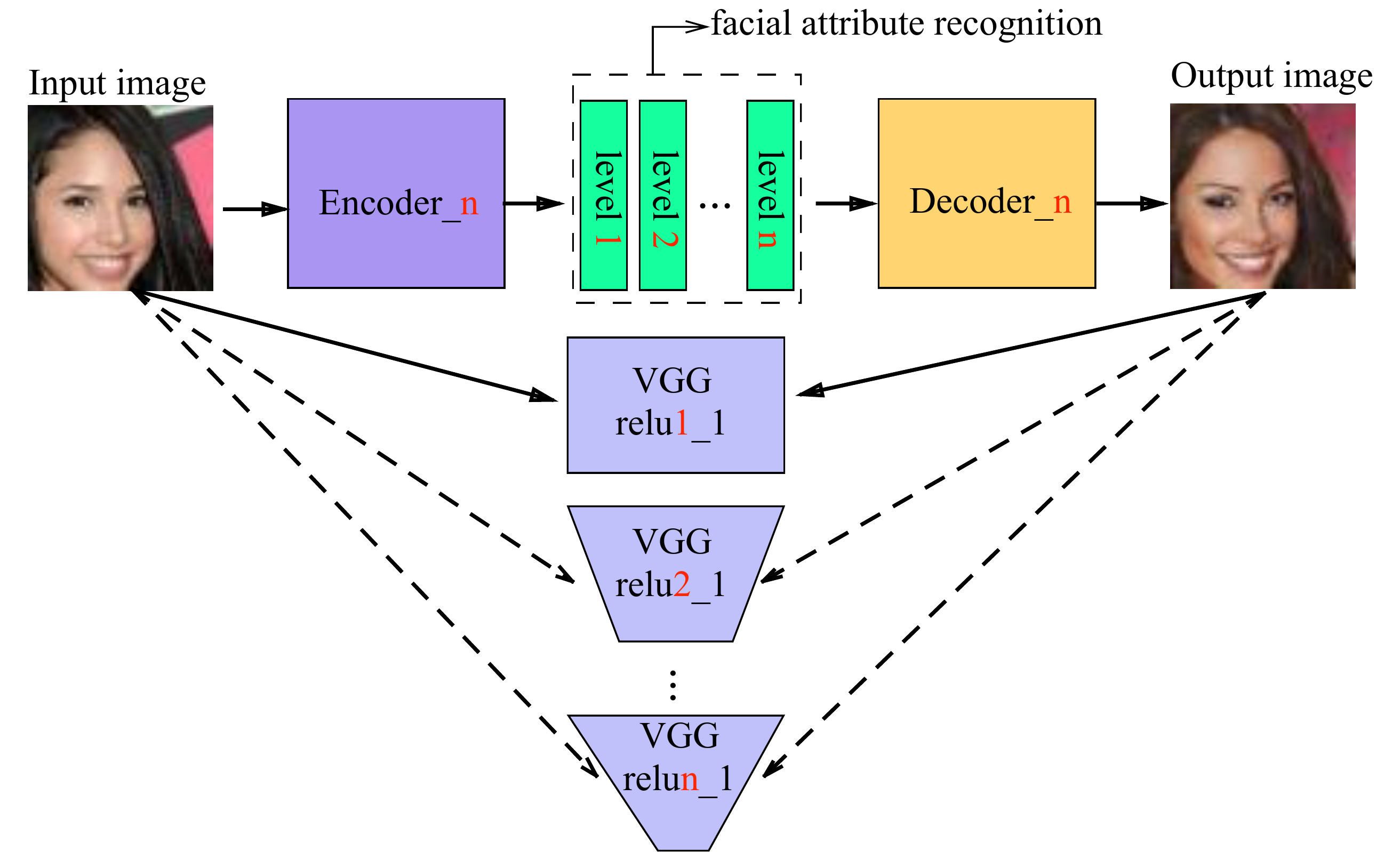}
\end{tabular}
\caption{Multi-view feature extraction. 5 VAE-WGAN models are trained with feature reconstruction loss based on layers relu1\_1, relu2\_1 relu3\_1, relu4\_1 and relu5\_1 respectively. The latent vectors for all the 5 models are concatenated as the final extracted features for facial attribute recognition.}
\label{fig:multi_features}
\end{figure*}

\begin{table}[!tb]
    \caption{Performance comparison of 40 facial attributes recognition.}
    \footnotesize
    \begin{center}
        \tabcolsep=0.08cm
        \begin{tabular}{  c  c  c  c  c  c  c  c  c  c  c  c  c  c  c  c  c  c  c  c  c  c  }
            \hline
            Method & \rotatebox[origin=c]{90}{5 Shadow} & \rotatebox[origin=c]{90}{Arch. Eyebrows} & \rotatebox[origin=c]{90} {Attractive} & \rotatebox[origin=c]{90} {Bags Un. Eyes} & \rotatebox[origin=c]{90} {Bald} & \rotatebox[origin=c]{90} {Bangs} & \rotatebox[origin=c]{90} {Big Lips} & \rotatebox[origin=c]{90} {Big Nose} & \rotatebox[origin=c]{90} {Black Hair} & \rotatebox[origin=c]{90} {Blond Hair} & \rotatebox[origin=c]{90} {Blurry} & \rotatebox[origin=c]{90} {Brown Hair} & \rotatebox[origin=c]{90} {Bushy Eyebrows} & \rotatebox[origin=c]{90} {Chubby} & \rotatebox[origin=c]{90} {Double Chin} & \rotatebox[origin=c]{90} {Eyeglasses} & \rotatebox[origin=c]{90} {Goatee} & \rotatebox[origin=c]{90} {Gray Hair} & \rotatebox[origin=c]{90} {Heavy Makeup} & \rotatebox[origin=c]{90} {H. Cheekbones} & \rotatebox[origin=c]{90} {Male} \\  \hline
            FaceTracer \citep{kumar2008facetracer}  & 85 & 76 & 78 & 76 & 89 & 88 & 64 & 74 & 70 & 80 & 81 & 60 & 80 & 86 & 88 & 98 & 93 & 90 & 85 & 84 & 91 \\ 
            PANDA-w \citep{zhang2014panda} & 82 & 73 & 77 & 71 & 92 & 89 & 61 & 70 & 74 & 81 & 77 & 69 & 76 & 82 & 85 & 94 & 86 & 88 & 84 & 80 & 93 \\ 
            PANDA-l \citep{zhang2014panda} & 88 & 78 & \textbf{81} & 79 & 96 & 92 & 67 & 75 & 85 & 93 & 86 & 77 & 86 & 86 & 88 & 98 & 93 & 94 & \textbf{90} & 86 & 97 \\ 
            LNets+ANet \citep{liu2015deep} & \textbf{91} & 79 & \textbf{81} & 79 & \textbf{98} & \textbf{95} & 68 & 78 & \textbf{88} & \textbf{95} & 84 & 80 & \textbf{90} & 91 & 92 & \textbf{99} & \textbf{95} & \textbf{97} & \textbf{90} & \textbf{87} & \textbf{98} \\ 
            VAE-123 \citep{hou2017deep} & 89 & 77 & 75 & 81 & \textbf{98} & 91 & 76 & 79 & 83 & 92 & \textbf{95} & 80 & 87 & 94 & 95 & 96 & 94 & 96 & 85 & 81 & 90 \\ 
            VAE-345 \citep{hou2017deep} & 89 &  \textbf{80} & 78 &  \textbf{82} &  \textbf{98} &  \textbf{95} &  \textbf{77} &  \textbf{81} & 85 & 93 &  \textbf{95} &  80 & 88 &  94 &  \textbf{96} &  \textbf{99} &  \textbf{95} &  \textbf{97} & 89 & 85 & 95 \\
            VGG-FC\citep{hou2017deep} & 83 & 71 & 68 & 73 & 97 & 81 & 51 & 77 & 78 & 88 & 94 & 67 & 81 & 93 & 93 & 95 & 93 & 94 & 79 & 64 & 84 \\ \hline
            VAE-WGAN (ours) & 90 & \textbf{80} & 79 & \textbf{82} & \textbf{98} & \textbf{95} & \textbf{77} & \textbf{81} & 86 & 94 & \textbf{95} & \textbf{82} & 89 & \textbf{95} & \textbf{96} & 98 & \textbf{95} & \textbf{97} & 88 & 85 & 94 \\  \hline
            \  & \  & \  & \  & \  & \  & \  & \  & \  & \  & \  & \  & \  & \  & \  & \  & \  & \  & \  & \  & \  & \  \\ \hline
            Method & \rotatebox[origin=c]{90}{Mouth S. O.} & \rotatebox[origin=c]{90}{Mustache} & \rotatebox[origin=c]{90}{Narrow Eyes} & \rotatebox[origin=c]{90}{No Beard} & \rotatebox[origin=c]{90}{Oval Face} & \rotatebox[origin=c]{90}{Pale Skin} & \rotatebox[origin=c]{90}{Pointy Nose} & \rotatebox[origin=c]{90}{Reced. Hairline} & \rotatebox[origin=c]{90}{Rosy Cheeks} & \rotatebox[origin=c]{90}{Sideburns} & \rotatebox[origin=c]{90}{Smiling} & \rotatebox[origin=c]{90}{Straight Hair} & \rotatebox[origin=c]{90}{Wavy Hair} & \rotatebox[origin=c]{90}{Wear. Earrings} & \rotatebox[origin=c]{90}{Wear. Hat} & \rotatebox[origin=c]{90}{Wear. Lipstick} & \rotatebox[origin=c]{90}{Wear. Necklace} & \rotatebox[origin=c]{90}{Wear. Necktie} & \rotatebox[origin=c]{90}{Young} & \rotatebox[origin=c]{90}{Average} & \  \\ \hline
            FaceTracer \citep{kumar2008facetracer} & 87 & 91 & 82 & 90 & 64 & 83 & 68 & 76 & 84 & 94 & 89 & 63 & 73 & 73 & 89 & 89 & 68 & 86 & 80 & 81.13 & \  \\ 
            PANDA-w \citep{zhang2014panda} & 82 & 83 & 79 & 87 & 62 & 84 & 65 & 82 & 81 & 90 & 89 & 67 & 76 & 72 & 91 & 88 & 67 & 88 & 77 & 79.85 & \  \\ 
            PANDA-l \citep{zhang2014panda} & \textbf{93} & 93 & 84 & 93 & 65 & 91 & 71 & 85 & 87 & 93 & \textbf{92} & 69 & 77 & 78 & 96 & \textbf{93} & 67 & 91 & 84 & 85.43 & \  \\ 
            LNets+ANet \citep{liu2015deep} & 92 & 95 & 81 & \textbf{95} & 66 & 91 & 72 & 89 & 90 & \textbf{96} & \textbf{92} & 73 & \textbf{80} & 82 & \textbf{99} & \textbf{93} & 71 & \textbf{93} & \textbf{87} & 87.30 & \  \\ 
            VAE-123 \citep{hou2017deep} & 80 & \textbf{96} & \textbf{89} & 88 & 73 & 96 & 73 & \textbf{92} & \textbf{94} & 95 & 87 & 79 & 74 & 82 & 96 & 88 & \textbf{88} & \textbf{93} & 81 & 86.95 & \  \\ 
            VAE-345 \citep{hou2017deep} & 88 &  \textbf{96} &  \textbf{89} & 91 &  \textbf{74} &  96 &  \textbf{74} &  \textbf{92} &  \textbf{94} &  \textbf{96} & 91 &  \textbf{80} & 79 &  84 & 98 & 91 &  \textbf{88} &  \textbf{93} & 84 &  88.73 & \  \\ 
            VGG-FC\citep{hou2017deep} & 60 & 93 & 87 & 84 & 66 &  96 & 58 & 86 & 93 & 85 & 65 & 68 & 70 & 49 & 98 & 82 & 87 & 89 & 74 & 79.85 & \  \\ \hline

            VAE-WGAN (ours) & 85 & \textbf{96} & \textbf{89} & 91 & \textbf{74} & \textbf{97} & \textbf{74} & \textbf{92} & \textbf{94} & \textbf{96} & 91 & \textbf{80} & \textbf{80} & \textbf{85} & \textbf{99} & 91 & \textbf{88} & \textbf{93} & 84 & \textbf{88.88} & \  \\ \hline
        \end{tabular}
    \end{center}
    \label{tab:performance_attribute_prediction}
\end{table}

\subsection{Facial attribute recognition}
We further evaluate the quality of the learned latent representations of the VAE by applying them to facial attribute recognition, which is a very challenging problem. Like \cite{liu2015deep}, 20,000 face images in the CelebA dataset \citep{liu2015deep} are used for testing while the remaining are used as training data. We proposed to use a multi-view strategy for feature extraction as shown in Figure \ref{fig:multi_features}. Specifically, 5 VAE-WGAN models are trained independently, each uses a different convolutional layer of the VGGNet to calculate the feature reconstruction loss. The latent vectors for all the 5 models are concatenated as the final extracted features which are used to train standard linear SVM classifiers to predict the 40 facial attributes in the dataset. As a result, we train 40 binary classifiers for each attribute in CelebA dataset respectively.

We then compare our method with other state of the art methods, i.e., FaceTracer \cite{kumar2008facetracer}, PANDA-w \cite{zhang2014panda}, PANDA-l \cite{zhang2014panda}, LNets+ANet \cite{liu2015deep},VAE-123 \citep{hou2017deep}, VAE-345 \citep{hou2017deep}. From Table \ref{tab:performance_attribute_prediction}, It is seen that our method can achieve the highest average prediction accuracies, which slightly beats the state of the art results. Additionally, we find that our method is not always the best for all the facial attributes. In particular, it does not work very well to predict attributes like “Mouth S. O” (mouth slightly open) and ``Wear Lipstick'' as shown in Table \ref{tab:performance_attribute_prediction}. One possible explanation of this is that these attributes are hard to detect in face images and difficult to reconstruct precisely in variational autoencoder model. As a result, the encoded latent vectors are not able to capture such subtle differences.


\section{Discussion}
\label{sec:disscussion}
For variational autoencoder model, one essential part is to define a metric to measure the inconsistency between the input and the reconstructed output. The plain VAE adopts the per-pixel measurement, leading to unacceptably blurry outputs because it essentially treats images as ``unstructured'' input and each pixel is independent with all the other pixels. Inspired by the recent works like image style transfer \cite{gatys2015neural,johnson2016perceptual,ulyanov2016texture}, we propose to improve the performance of VAE by measuring the inconsistency in the deep feature space instead of naive pixel space. The hidden representations from pretrained deep CNN are able to capture essential visual quality factors such as spatial correlation because of convolutional operations. What's more, variational autoencoder can be seamlessly incorporated into the framework of generative adversarial network to enforce the output to resemble natural images. The adversarial loss can be regarded as ``structured'' measurement because the GAN training is essentially performing high level image classification, and each pixel is not treated independently at all.

Another benefit of using deep CNNs to construct loss function is that we can achieve multi-scale modeling implicitly. Due to the hierarchy architecture of deep convolutional neural networks, a higher layer is corresponding to a coarser spatial area of the encoded image. Thus, unlike traditional methods that try to directly use multi-scale images as input, we can achieve another kind of multi-scale modeling by constructing loss function with different layers.

Another interesting part of VAE is the linear property in the learned latent space. Different images generated by the decoder can be smoothly transformed to others by a simple linear combination of their latent vectors. Additionally attribute specific features could be also calculated by encoding the annotated images and used to manipulate the related attribute of a given image while keeping other attributes unchanged.

\section{Conclusion}
\label{sec:conclusion}
In this paper, we propose a more stable architecture and several effective techniques to incorporate variational autoencoder. In particular, we employ deep feature consistent principle to allow the output to have a better perceptual quality and use adversarial training to help produce images that reside on the manifold of natural images. Compared to previous approaches, our model can generate more consistent and realistic images with fine details and reasonable backgrounds. In addition, we further investigate the quality of the learned representation to manipulate facial attributes. Finally, we have shown that our method can be used to extract effective representations for facial attribute recognition and achieve state of the art performance.

\section*{References}

\bibliography{mybibfile}

\begin{thebibliography}{10}
\expandafter\ifx\csname url\endcsname\relax
  \def\url#1{\texttt{#1}}\fi
\expandafter\ifx\csname urlprefix\endcsname\relax\def\urlprefix{URL }\fi
\expandafter\ifx\csname href\endcsname\relax
  \def\href#1#2{#2} \def\path#1{#1}\fi

\bibitem{GU2018354}
J.~Gu, Z.~Wang, J.~Kuen, L.~Ma, A.~Shahroudy, B.~Shuai, T.~Liu, X.~Wang,
  G.~Wang, J.~Cai, T.~Chen, Recent advances in convolutional neural networks,
  Pattern Recognition 77 (2018) 354 -- 377.

\bibitem{krizhevsky2012imagenet}
A.~Krizhevsky, I.~Sutskever, G.~E. Hinton, Imagenet classification with deep
  convolutional neural networks, in: Advances in neural information processing
  systems, 2012, pp. 1097--1105.

\bibitem{simonyan2014very}
K.~Simonyan, A.~Zisserman, Very deep convolutional networks for large-scale
  image recognition, International Conference on Learning Representations
  (ICLR), 2015.

\bibitem{he2016deep}
K.~He, X.~Zhang, S.~Ren, J.~Sun, Deep residual learning for image recognition,
  in: Proceedings of the IEEE conference on computer vision and pattern
  recognition, 2016, pp. 770--778.

\bibitem{babenko2014neural}
A.~Babenko, A.~Slesarev, A.~Chigorin, V.~Lempitsky, Neural codes for image
  retrieval, in: Computer Vision--ECCV 2014, Springer, 2014, pp. 584--599.

\bibitem{girshick2014rich}
R.~Girshick, J.~Donahue, T.~Darrell, J.~Malik, Rich feature hierarchies for
  accurate object detection and semantic segmentation, in: Proceedings of the
  IEEE conference on computer vision and pattern recognition, 2014, pp.
  580--587.

\bibitem{karpathy2015deep}
A.~Karpathy, L.~Fei-Fei, Deep visual-semantic alignments for generating image
  descriptions, in: Proceedings of the IEEE Conference on Computer Vision and
  Pattern Recognition, 2015, pp. 3128--3137.

\bibitem{yu2018multitask}
J.~Yu, C.~Hong, Y.~Rui, D.~Tao, Multitask autoencoder model for recovering
  human poses, IEEE Transactions on Industrial Electronics 65~(6) (2018)
  5060--5068.

\bibitem{hong2016hypergraph}
C.~Hong, X.~Chen, X.~Wang, C.~Tang, Hypergraph regularized autoencoder for
  image-based 3d human pose recovery, Signal Processing 124 (2016) 132--140.

\bibitem{hong2015image}
C.~Hong, J.~Yu, D.~Tao, M.~Wang, Image-based three-dimensional human pose
  recovery by multiview locality-sensitive sparse retrieval, IEEE Transactions
  on Industrial Electronics 62~(6) (2015) 3742--3751.

\bibitem{yu2017iprivacy}
J.~Yu, B.~Zhang, Z.~Kuang, D.~Lin, J.~Fan, iprivacy: image privacy protection
  by identifying sensitive objects via deep multi-task learning, IEEE
  Transactions on Information Forensics and Security 12~(5) (2017) 1005--1016.

\bibitem{zhang2018local}
J.~Zhang, J.~Yu, D.~Tao, Local deep-feature alignment for unsupervised
  dimension reduction, IEEE Transactions on Image Processing 27~(5) (2018)
  2420--2432.

\bibitem{liu2018balance}
A.~Liu, Y.~Laili, Balance gate controlled deep neural network, Neurocomputing
  320 (2018) 183--194.

\bibitem{junior2018randomized}
J.~J. de~Mesquita Sá~Junior, A.~R. Backes, O.~M. Bruno, Randomized neural
  network based descriptors for shape classification, Neurocomputing 312 (2018)
  201 -- 209.

\bibitem{osipov2018space}
V.~Osipov, M.~Osipova, Space–time signal binding in recurrent neural networks
  with controlled elements, Neurocomputing 308 (2018) 194 -- 204.

\bibitem{kingma2013auto}
D.~P. Kingma, M.~Welling, Auto-encoding variational bayes, International
  Conference on Learning Representations (ICLR), 2014.

\bibitem{hou2017deep}
X.~Hou, L.~Shen, K.~Sun, G.~Qiu, Deep feature consistent variational
  autoencoder, in: Applications of Computer Vision (WACV), 2017 IEEE Winter
  Conference on, IEEE, 2017, pp. 1133--1141.

\bibitem{goodfellow2014generative}
I.~Goodfellow, J.~Pouget-Abadie, M.~Mirza, B.~Xu, D.~Warde-Farley, S.~Ozair,
  A.~Courville, Y.~Bengio, Generative adversarial nets, in: Advances in Neural
  Information Processing Systems, 2014, pp. 2672--2680.

\bibitem{arjovsky2017wasserstein}
M.~Arjovsky, S.~Chintala, L.~Bottou, Wasserstein generative adversarial
  networks, in: Proceedings of the 34th International Conference on Machine
  Learning, {ICML} 2017, Sydney, NSW, Australia, 6-11 August 2017, 2017, pp.
  214--223.

\bibitem{makkie2019fast}
M.~Makkie, H.~Huang, Y.~Zhao, A.~V. Vasilakos, T.~Liu, Fast and scalable
  distributed deep convolutional autoencoder for fmri big data analytics,
  Neurocomputing 325 (2019) 20--30.

\bibitem{chen2018cross}
J.~Chen, Z.~Wu, J.~Zhang, F.~Li, W.~Li, Z.~Wu, Cross-covariance regularized
  autoencoders for nonredundant sparse feature representation, Neurocomputing
  316 (2018) 49--58.

\bibitem{chen2018evolutionary}
Z.~Chen, C.~K. Yeo, B.~S. Lee, C.~T. Lau, Y.~Jin, Evolutionary multi-objective
  optimization based ensemble autoencoders for image outlier detection,
  Neurocomputing.

\bibitem{feng2018graph}
S.~Feng, M.~F. Duarte, Graph autoencoder-based unsupervised feature selection
  with broad and local data structure preservation, Neurocomputing 312 (2018)
  310 -- 323.

\bibitem{sun2017generalized}
K.~Sun, J.~Zhang, C.~Zhang, J.~Hu, Generalized extreme learning machine
  autoencoder and a new deep neural network, Neurocomputing 230 (2017)
  374--381.

\bibitem{hong2015multimodal}
C.~Hong, J.~Yu, J.~Wan, D.~Tao, M.~Wang, Multimodal deep autoencoder for human
  pose recovery, IEEE Transactions on Image Processing 24~(12) (2015)
  5659--5670.

\bibitem{rezende2014stochastic}
D.~J. Rezende, S.~Mohamed, D.~Wierstra, Stochastic backpropagation and
  approximate inference in deep generative models, in: Proceedings of the 31st
  International Conference on Machine Learning (ICML-14), 2014, pp. 1278--1286.

\bibitem{kingma2014semi}
D.~P. Kingma, S.~Mohamed, D.~J. Rezende, M.~Welling, Semi-supervised learning
  with deep generative models, in: Advances in Neural Information Processing
  Systems, 2014, pp. 3581--3589.

\bibitem{yan2015attribute2image}
X.~Yan, J.~Yang, K.~Sohn, H.~Lee, Attribute2image: Conditional image generation
  from visual attributes, in: Computer Vision - {ECCV} 2016 - 14th European
  Conference, Amsterdam, The Netherlands, October 11-14, 2016, Proceedings,
  Part {IV}, 2016, pp. 776--791.

\bibitem{gregor2015draw}
K.~Gregor, I.~Danihelka, A.~Graves, D.~J. Rezende, D.~Wierstra, {DRAW:} {A}
  recurrent neural network for image generation, in: {ICML}, Vol.~37 of {JMLR}
  Workshop and Conference Proceedings, JMLR.org, 2015, pp. 1462--1471.

\bibitem{ridgeway2015learning}
K.~Ridgeway, J.~Snell, B.~Roads, R.~Zemel, M.~Mozer, Learning to generate
  images with perceptual similarity metrics, arXiv preprint arXiv:1511.06409.

\bibitem{denton2015deep}
E.~L. Denton, S.~Chintala, R.~Fergus, et~al., Deep generative image models
  using a laplacian pyramid of adversarial networks, in: Advances in neural
  information processing systems, 2015, pp. 1486--1494.

\bibitem{radford2015unsupervised}
A.~Radford, L.~Metz, S.~Chintala, Unsupervised representation learning with
  deep convolutional generative adversarial networks, arXiv preprint
  arXiv:1511.06434.

\bibitem{im2016generating}
D.~J. Im, C.~D. Kim, H.~Jiang, R.~Memisevic, Generating images with recurrent
  adversarial networks, arXiv preprint arXiv:1602.05110.

\bibitem{salimans2016improved}
T.~Salimans, I.~J. Goodfellow, W.~Zaremba, V.~Cheung, A.~Radford, X.~Chen,
  Improved techniques for training gans, in: Advances in Neural Information
  Processing Systems 29: Annual Conference on Neural Information Processing
  Systems 2016, December 5-10, 2016, Barcelona, Spain, 2016, pp. 2226--2234.

\bibitem{chen2016infogan}
X.~Chen, Y.~Duan, R.~Houthooft, J.~Schulman, I.~Sutskever, P.~Abbeel, Infogan:
  Interpretable representation learning by information maximizing generative
  adversarial nets, in: Advances in Neural Information Processing Systems 29:
  Annual Conference on Neural Information Processing Systems 2016, December
  5-10, 2016, Barcelona, Spain, 2016, pp. 2172--2180.

\bibitem{karras2018progressive}
T.~{Karras}, T.~{Aila}, S.~{Laine}, J.~{Lehtinen}, Progressive growing of gans
  for improved quality, stability, and variation, International Conference on
  Learning Representations (ICLR), 2018.

\bibitem{hu2018unifying}
Z.~{Hu}, Z.~{Yang}, R.~{Salakhutdinov}, E.~P. {Xing}, On unifying deep
  generative models, International Conference on Learning Representations
  (ICLR), 2018.

\bibitem{dong2018san}
X.~Dong, Y.~Yan, W.~Ouyang, Y.~Yang, Style aggregated network for facial
  landmark detection, in: Proceedings of the IEEE Conference on Computer Vision
  and Pattern Recognition, 2018.

\bibitem{DBLP:conf/cvpr/DongHYY17}
X.~Dong, J.~Huang, Y.~Yang, S.~Yan, More is less: {A} more complicated network
  with less inference complexity, in: 2017 {IEEE} Conference on Computer Vision
  and Pattern Recognition, {CVPR} 2017, Honolulu, HI, USA, July 21-26, 2017,
  {IEEE} Computer Society, 2017, pp. 1895--1903.

\bibitem{DBLP:conf/cvpr/WanPGY17}
C.~Wan, T.~Probst, L.~V. Gool, A.~Yao, Crossing nets: Combining gans and vaes
  with a shared latent space for hand pose estimation, in: 2017 {IEEE}
  Conference on Computer Vision and Pattern Recognition, {CVPR} 2017, Honolulu,
  HI, USA, July 21-26, 2017, {IEEE} Computer Society, 2017, pp. 1196--1205.

\bibitem{DBLP:journals/corr/Qi17}
G.~Qi, Loss-sensitive generative adversarial networks on lipschitz densities,
  CoRR abs/1701.06264.
\newblock \href {http://arxiv.org/abs/1701.06264} {\path{arXiv:1701.06264}}.

\bibitem{larsen2015autoencoding}
A.~B.~L. Larsen, S.~K. S{\o}nderby, H.~Larochelle, O.~Winther, Autoencoding
  beyond pixels using a learned similarity metric, in: Proceedings of the 33nd
  International Conference on Machine Learning, {ICML} 2016, New York City, NY,
  USA, June 19-24, 2016, 2016, pp. 1558--1566.

\bibitem{gatys2015neural}
L.~A. Gatys, A.~S. Ecker, M.~Bethge, A neural algorithm of artistic style,
  arXiv preprint arXiv:1508.06576.

\bibitem{johnson2016perceptual}
J.~Johnson, A.~Alahi, L.~Fei{-}Fei, Perceptual losses for real-time style
  transfer and super-resolution, in: Computer Vision - {ECCV} 2016 - 14th
  European Conference, Amsterdam, The Netherlands, October 11-14, 2016,
  Proceedings, Part {II}, 2016, pp. 694--711.

\bibitem{ulyanov2016texture}
D.~Ulyanov, V.~Lebedev, A.~Vedaldi, V.~S. Lempitsky, Texture networks:
  Feed-forward synthesis of textures and stylized images, in: Proceedings of
  the 33nd International Conference on Machine Learning, {ICML} 2016, New York
  City, NY, USA, June 19-24, 2016, 2016, pp. 1349--1357.

\bibitem{li2016combining}
C.~Li, M.~Wand, Combining markov random fields and convolutional neural
  networks for image synthesis, in: 2016 {IEEE} Conference on Computer Vision
  and Pattern Recognition, {CVPR} 2016, Las Vegas, NV, USA, June 27-30, 2016,
  2016, pp. 2479--2486.

\bibitem{simonyan2013deep}
K.~Simonyan, A.~Vedaldi, A.~Zisserman, Deep inside convolutional networks:
  Visualising image classification models and saliency maps, arXiv preprint
  arXiv:1312.6034.

\bibitem{yosinski2015understanding}
J.~Yosinski, J.~Clune, A.~Nguyen, T.~Fuchs, H.~Lipson, Understanding neural
  networks through deep visualization, arXiv preprint arXiv:1506.06579.

\bibitem{szegedy2013intriguing}
C.~{Szegedy}, W.~{Zaremba}, I.~{Sutskever}, J.~{Bruna}, D.~{Erhan}, I.~J.
  {Goodfellow}, R.~{Fergus}, Intriguing properties of neural networks,
  International Conference on Learning Representations (ICLR), 2014.

\bibitem{nguyen2015deep}
A.~Nguyen, J.~Yosinski, J.~Clune, Deep neural networks are easily fooled: High
  confidence predictions for unrecognizable images, in: Proceedings of the IEEE
  Conference on Computer Vision and Pattern Recognition, 2015, pp. 427--436.

\bibitem{long2015fully}
J.~Long, E.~Shelhamer, T.~Darrell, Fully convolutional networks for semantic
  segmentation, in: Proceedings of the IEEE Conference on Computer Vision and
  Pattern Recognition, 2015, pp. 3431--3440.

\bibitem{liu2015deep}
Z.~Liu, P.~Luo, X.~Wang, X.~Tang, Deep learning face attributes in the wild,
  in: Proceedings of the IEEE International Conference on Computer Vision,
  2015, pp. 3730--3738.

\bibitem{Krizhevsky09}
A.~Krizhevsky, G.~Hinton, Learning multiple layers of features from tiny
  images, Master's thesis, Department of Computer Science, University of
  Toronto.

\bibitem{kingma2014adam}
D.~P. {Kingma}, J.~L. {Ba}, Adam: A method for stochastic optimization,
  International Conference on Learning Representations (ICLR), 2015.

\bibitem{collobert2011torch7}
R.~Collobert, K.~Kavukcuoglu, C.~Farabet, Torch7: A matlab-like environment for
  machine learning, in: BigLearn, NIPS Workshop, no. EPFL-CONF-192376, 2011.

\bibitem{mikolov2013distributed}
T.~Mikolov, J.~Dean, Distributed representations of words and phrases and their
  compositionality, Advances in neural information processing systems, 2013.

\bibitem{maaten2008visualizing}
L.~v.~d. Maaten, G.~Hinton, Visualizing data using t-sne, Journal of Machine
  Learning Research 9~(Nov) (2008) 2579--2605.

\bibitem{kumar2008facetracer}
N.~Kumar, P.~Belhumeur, S.~Nayar, Facetracer: A search engine for large
  collections of images with faces, in: European conference on computer vision,
  Springer, 2008, pp. 340--353.

\bibitem{zhang2014panda}
N.~Zhang, M.~Paluri, M.~Ranzato, T.~Darrell, L.~Bourdev, Panda: Pose aligned
  networks for deep attribute modeling, in: Proceedings of the IEEE Conference
  on Computer Vision and Pattern Recognition, 2014, pp. 1637--1644.

\end{thebibliography}

\end{document}